\documentclass{article}

\usepackage[preprint]{neurips_2025}
\workshoptitle{AI for Science}

\usepackage[utf8]{inputenc} 
\usepackage[T1]{fontenc}    
\usepackage{hyperref}       
\usepackage{url}            
\usepackage{booktabs}       
\usepackage{amsfonts}       
\usepackage{nicefrac}       
\usepackage{microtype}      
\usepackage{xcolor}        
\usepackage{amsmath}
\usepackage{mathtools}
\usepackage{tikz}
\usetikzlibrary{positioning, shapes.geometric, arrows.meta, fit, backgrounds}
\tikzset{
    block/.style={draw, thick, minimum width=2.2cm, minimum height=0.6cm, align=center, rounded corners, fill=#1, font=\scriptsize},
    arrow/.style={-{Latex[length=1.5mm]}, thick},
    group/.style={draw, thick, rounded corners, inner sep=4pt}
}
\usepackage{float}
\usepackage{wrapfig}
\usepackage{multirow}
\usepackage{graphicx}
\usepackage{caption}
\usepackage{subcaption}
\usepackage{geometry}
\usetikzlibrary{calc}

\title{Physics-Informed Neural Networks with Fourier Features and Attention-Driven Decoding}

\author{ 
  Rohan Arni \thanks{Alternate email: rohan.arni@gmail.com} \\
  High Technology High School\\
  Lincroft, NJ 07738 \\
  \texttt{roarni@ctemc.org} \\
  \And
  Carlos Blanco \\
  Institute for Gravitation and the Cosmos \\
  The Pennsylvania State University \\
  University Park, PA 16802 \\
  \&\\
  Department of Physics\\
  Princeton University\\
  Princeton, NJ, 08544\\
  \texttt{carlosblanco@psu.edu} \\
}

\begin{document}

\maketitle

\begin{abstract}

Physics-Informed Neural Networks (PINNs) are a useful framework for approximating partial differential equation solutions using deep learning methods. In this paper, we propose a principled redesign of the PINNsformer, a Transformer-based PINN architecture. We present the Spectral PINNSformer (S-Pformer), a refinement of encoder-decoder PINNSformers that addresses two key issues; 1. the redundancy (i.e. increased parameter count) of the encoder, and 2. the mitigation of spectral bias. We find that the encoder is unnecessary for capturing spatiotemporal correlations when relying solely on self-attention, thereby reducing parameter count. Further, we integrate Fourier feature embeddings to explicitly mitigate spectral bias, enabling adaptive encoding of multiscale behaviors in the frequency domain. Our model outperforms encoder-decoder PINNSformer architectures across all benchmarks, achieving or outperforming MLP performance while reducing parameter count significantly. 
\end{abstract}

\section{Introduction}

Numerically computing solutions to partial differential equations has been a key area of research in science and engineering. Typical computational methods such as the finite difference method~\citep{NAKAYAMA2018293} or the spectral method suffer~\citep{Fornberg_1996} from computational overhead due to fine spatial-temporal discretization. These grid-based approaches also struggle with adaptability to complex geometries and require careful treatment of boundary conditions, which can increase the computational cost and implementation complexity~\citep{vmurugesh_2025_a}.

In recent years, Physics-Informed Neural Networks (PINNs) have emerged as a promising alternative, as they utilize deep learning methods to approximate the solution of PDEs by embedding the governing physical laws directly into the loss function~\citep{RAISSI2019686}. Rather than explicitly discretizing the domain, PINNs use neural networks that accept continuous space-time coordinates as input, enabling mesh-free solution approximations.

The majority of PINNs rely on multilayer perceptrons (MLPs), which suffer from spectral bias: a limitation where the networks have difficulty learning high-frequency components in differential equation solutions~\citep{wangspectralbias}. In addition, MLP-based PINNs can suffer from limited generalization when applied to more challenging or nonlinear problems due to simplicity bias~\citep{xu2025subsequentialphysicsinformedlearningstate}.

To address these limitations, attention-based architectures such as the PINNsformer have been developed~\citep{zhao2024pinnsformertransformerbasedframeworkphysicsinformed}. These models use the self-attention mechanism from Transformers, allowing the network to attend to relevant spatial and temporal relationships in the input. The attention mechanism allows for dramatic improvements in performance. 

In this work, we propose a Spectral PINNsformer (S-Pformer) architecture that streamlines the architecture by replacing the encoder layer with Fourier feature embeddings and applying self-attention directly within the decoder. The Fourier feature embeddings enabling the S-Pformer to better capture multiscale behaviors by adaptively projecting input coordinates into a spectral representation in order to better understand different frequencies in the PDE solution~\citep{tancik2020fourierfeaturesletnetworks}. In addition, replacing the encoder helps reduce parameter count while still maintaining high performance on PDE benchmarks.

\section{Methodology}

\subsection{Background}

Let $d$ be the number of input spatial dimensions. Let $\Omega$ be an open set in $\mathbb{R}^{d}$ bounded by $\partial\Omega$. A PDE with spatiotemporal input $(\mathbf{x}, t)$ where $\mathbf{x} \in \mathbb{R}^d$ and $t \in \mathbb{R}$ follows the abstraction:

\begin{align}
\mathcal{F}(u(\mathbf{x}, t)) &= 0, && \forall(\mathbf{x}, t) \in \Omega \times [0, T], \\
\mathcal{I}(u(\mathbf{x}, 0)) &= 0, && \forall \mathbf{x} \in \Omega, \\
\mathcal{B}(u(\mathbf{x}, t)) &= 0, && \forall(\mathbf{x}, t) \in \partial\Omega \times [0, T]
\end{align}

where \( u: \mathbb{R}^{d+1} \rightarrow \mathbb{R}^m \) is the solution to the PDE. The operator \( \mathcal{F} \) encodes the physical law (the PDE in residual form), \( \mathcal{I} \) represents initial conditions, and \( \mathcal{B} \) encodes boundary conditions. Physics-Informed Neural Networks (PINNs) approximate \( u(\mathbf{x}, t) \) by enforcing these constraints during training through the loss function:

\begin{align} \label{eq:lossfn}
    \mathcal{L}(u_\theta) = 
    &\frac{\lambda_1}{N_\mathcal{F}} \sum_{i=1}^{N_\mathcal{F}} \left\| \mathcal{F}(u_\theta(\mathbf{x}_i, t_i)) \right\|^2 \notag \\
    &+  \frac{\lambda_2}{N_\mathcal{I}} \sum_{i=1}^{N_\mathcal{I}} \left\| \mathcal{I}(u_\theta(\mathbf{x}_i, 0)) \right\|^2 \notag \\
    &+ \frac{\lambda_3}{N_\mathcal{B}} \sum_{i=1}^{N_\mathcal{B}} \left\| \mathcal{B}(u_\theta(\mathbf{x}_i, t_i)) \right\|^2
\end{align}

where $u_\theta$ is the neural network approximation of the PDE solution, \( N_\mathcal{F}, N_\mathcal{I}, N_\mathcal{B} \) are the number of training points used for PDE, initial, and boundary regions, and \( \lambda_1, \lambda_2, \lambda_3 \) are respective weighting coefficients for each loss term. 

Traditional PINNs focus on point-wise predictions without incorporating temporal dependencies in PDE solutions. This approach is suitable primarily for elliptic PDEs, which lack explicit time derivatives. However, hyperbolic and parabolic PDEs involve time derivatives, which the architecture of MLP-based PINNs does not inherently account for~\citep{zhao2024pinnsformertransformerbasedframeworkphysicsinformed}.

\subsection{Previous Architecture}

To address the lack of spatiotemporal relationships in MLP-based PINNs, the PINNsformer architecture was proposed. PINNsformers, which are based on encoder-decoder transformer architectures, rely on attention mechanisms to better capture spatio-temporal relationships compared to a traditional MLP PINN. The PINNsformer architecture is built on four main components: a pseudo-sequence generator, a spatio-temporal mixer, an encoder-decoder with multi-head attention, and an output layer~\citep{zhao2024pinnsformertransformerbasedframeworkphysicsinformed}.

\paragraph{Pseudo-Sequence Generator}
Transformer-based models are trained on sequential data that are not compatible with the singular points used to train MLP-based PINNs. To account for this, the PINNsformer paper proposed the pseudo-sequence generator~\citep{zhao2024pinnsformertransformerbasedframeworkphysicsinformed}, which performs the following operation given a single space-time coordinate:

\begin{equation}
(\mathbf{x}, t) \xRightarrow{\text{Gen}} \{(\mathbf{x}, t), (\mathbf{x}, t + \Delta t), \ldots, (\mathbf{x}, t + (k - 1)\Delta t)\}
\end{equation}

where $\Delta t$ is a very small constant, and $k$ is the number of timesteps. This creates a temporal sequence while preserving the underlying physical relationships ~\citep{zhao2024pinnsformertransformerbasedframeworkphysicsinformed}. In practice, this means taking a single spatiotemporal coordinate and generating a temporal sequence by holding the spatial location $\mathbf{x}$ constant while creating $k$ time points separated by intervals of $\Delta t$.

\paragraph{Encoder-Decoder Architecture} The Encoder-Decoder, inspired by Transformers, includes multiple layers combining self-attention and feedforward operations in the encoder, while the decoder excludes self-attention and reuses the encoder’s embeddings~\citep{vaswani2023attentionneed}.  This design enables effective spatio-temporal dependency learning in differential equation solutions. ~\citep{zhao2024pinnsformertransformerbasedframeworkphysicsinformed}.

\paragraph{Wavelet Activation Function}The Wavelet Activation function is introduced in place of traditional non-linear functions like ReLU and LayerNorm, which can be suboptimal for Physics-Informed Neural Networks (PINNs) due to issues like discontinuous derivatives~\citep{zhao2024pinnsformertransformerbasedframeworkphysicsinformed}. Inspired by the Real Fourier Transform, the Wavelet function captures periodic behavior effectively without requiring prior knowledge of the solution. This is formulated as a weighted sum of sine and cosine:

\begin{equation}
    \text{Wavelet}(z) = \omega_1 \sin(z) + \omega_2 \cos(z)
\end{equation}

where $\omega_1$ and $\omega_2$ are learnable parameters. The Wavelet Activation function is used in the PINNsformer architecture as an activation function in the encoder and decoder~\citep{zhao2024pinnsformertransformerbasedframeworkphysicsinformed}.

\subsection{Spectral Architecture}

The encoder in the PINNsformer potentially introduces unnecessary computational overhead and parameter redundancy. Traditional encoder-decoder architectures were designed for sequence-to-sequence tasks where input and output have different structures (e.g., translation)~\citep{gao2022encoderdecoderredundantneuralmachine}. The encoder creates representations that the decoder must then reinterpret, creating an potentially unnecessary computational bottleneck. This two-stage processing adds complexity without a potentially corresponding benefit. Furthermore, the encoder does not directly address spectral bias, a fundamental limitation in neural PDE solvers~\citep{wangspectralbias}.

\begin{figure}[h]
    \vspace{0.5em}
    \centering
    \begin{tikzpicture}[node distance=0.7cm and 1.2cm, scale=0.7, transform shape]
        \node[block=red!10] (fourier) {Fourier Embedding};
        \node[circle, draw, thick, inner sep=1.5pt, right=of fourier] (sum) {$+$};
        \node[block=red!10, right=of sum] (pos) {Positional Embedding};

        \node[block=yellow!30, below=of sum] (mha1) {Multi-Head Attention};
        \node[block=orange!20, below=of mha1] (add1) {Add \& Wavelet};
        \node[block=blue!20, below=of add1] (ffn1) {Feed Forward};
        \node[block=orange!20, below=of ffn1] (add2) {Add \& Wavelet};
        \node[block=green!20, below= 0.6cm of add2] (mlp) {MLP};
        \node[block=gray!20, below=0.5cm of mlp] (output) {Decoder Output};

        \node[group, fit=(mha1)(add1)(ffn1)(add2), inner ysep=0.65em] (decoderlayer) {};
        \node[group, fit=(decoderlayer)(mlp), inner ysep=0.4em] (decoder) {};

        \draw[arrow] (fourier) -- (sum);
        \draw[arrow] (pos) -- (sum);
        \draw[arrow] (sum) -- (mha1);
        \draw[arrow] (mha1) -- (add1);
        \draw[arrow] (add1) -- (ffn1);
        \draw[arrow] (ffn1) -- (add2);
        \draw[arrow] (add2) -- (mlp);
        \draw[arrow] (mlp) -- (output);
    \end{tikzpicture}
    \caption{Spectral PINNsformer Architecture}
    \label{fig:decoder_parallel_sum}
    \vspace{-2.0em}
\end{figure}
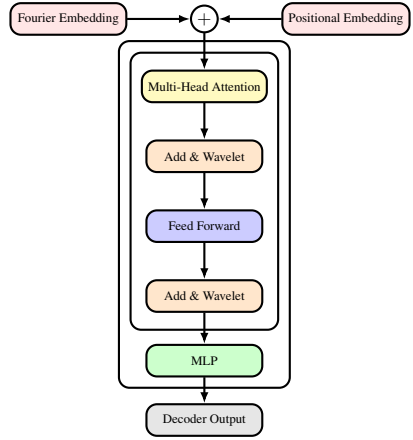

Therefore, we propose a Spectral PINNsformer (S-Pformer), which is a decoder-only transformer architecture for efficiency and simplicity. This new architecture (shown in Figure~\ref{fig:decoder_parallel_sum})is smaller than the full PINNsformer architecture, inherently prevents spectral bias with fourier features, and offers improved performance compared to previous architectures. The S-Pformer consists of three parts: Input embeddings with Fourier features, a decoder with multi-head attention, and an output linear layer. 

\subsubsection{Embeddings}

The input, given as a sequence of space-time points from the pseudo-sequence generator, is encoded using the modified embeddings module. It combines two components: a Fourier feature mapping and a learnable positional embedding. The Fourier mapping transforms the low-dimensional input $[\mathbf{x}, t]$ into a higher-dimensional periodic space using sine and cosine functions applied to randomized frequency projections~\citep{tancik2020fourierfeaturesletnetworks}. The Fourier feature mapping allows the model to inherently resolve spectral bias~\citep{wangspectralbias}. In parallel, the positional embedding applies a linear transformation directly to the input coordinates, which preserves spatial and temporal locality. This feature mapping directly replaces the spatio-temporal mixer and the encoder from the original architecture. Fourier features encode global periodic patterns essential for capturing oscillatory PDE solutions across multiple frequency scales, while positional embeddings preserve local spatial-temporal relationships.

Let $\mathbf{z} = (\mathbf{x}, t) \in \mathbb{R}^{d_\text{in}}$ denote the input coordinate where $d_\text{in} = d + 1$ (spatial dimensions plus time). The model first normalizes the coordinate to $\tilde{\mathbf{z}} = (\tilde{\mathbf{x}}, \tilde{t}) \in [0,1]^{d_\text{in}}$ such that all components fall within the range $[0, 1]$. 

This normalized input is then fed into our embedding module with output dimension $d_\text{emb}$. It consists of two parts: a Fourier feature embedding, and a positional embedding. The Fourier embedding depends on $d_\text{mapping}$, which determines the number of frequency bands that the input is projected into. We use a random projection matrix $\mathbf{B} \in \mathbb{R}^{d_\text{mapping} \times d_\text{in}}$ sampled from $\mathcal{N}(0, \sigma^2\mathbf{I})$. Since the input coordinates are normalized to [0,1], we set $\sigma^2= 1$  to ensure the random frequency projections have unit variance. Given a linear transformation $\theta_f: \mathbb{R}^{2d_\text{mapping}} \to \mathbb{R}^{d_\text{emb}}$, the Fourier feature embedding $E_f(\tilde{\mathbf{z}})$ is:

\begin{equation}
    E_f(\mathbf{\tilde{z}}) = \theta_f \left(\begin{bmatrix} \sin(2\pi\mathbf{B} \mathbf{\tilde{z}}) \\ \cos(2\pi\mathbf{B} \mathbf{\tilde{z}}) \end{bmatrix}\right)
\end{equation}

The positional embedding applies a linear transformation $\theta_p: \mathbb{R}^{d_\text{in}} \to \mathbb{R}^{d_\text{emb}}$ to preserve spatial-temporal locality:

\begin{equation}
    E_p(\mathbf{\tilde{z}}) = \theta_p(\mathbf{\tilde{z}})
\end{equation}

Finally, the combined input embedding is written as

\begin{equation}
    E(\mathbf{\tilde{z}}) = E_f(\mathbf{\tilde{z}}) + E_p(\mathbf{\tilde{z}})
\end{equation}

\subsubsection{Attention-Driven Decoder}

The decoder processes the embedded input through $N$ transformer layers, where each layer consists of multi-head attention followed by a feed-forward network, with residual connections and wavelet activations.

Let $\phi(z)$ denote the Wavelet activation function. Let $\mathrm{MH}(Q, K, V)$ denote a multi-head attention operation with query $Q$, key $K$, and value $V$ matrices. The feed-forward component $\mathrm{FF}(x)$ is defined as a 3-layer MLP  with hidden dimension $d_\text{ff}$.

The decoder processes $N$ layers sequentially. We initialize with the embedding output:
\begin{equation}
    H^{(0)} = E(\mathbf{\tilde{z}})
\end{equation}

Each decoder layer applies the following transformations:

\begin{align}\label{eq:decoder_transform}
    U^{(l)} &= \phi(H^{(l-1)}) \quad \text{(Pre-attention normalization)} \\
    A^{(l)} &= \mathrm{MH}(U^{(l)}, U^{(l)}, U^{(l)}) \quad \text{(Self-attention)} \\
    S^{(l)} &= H^{(l-1)} + A^{(l)} \quad \text{(Residual connection)} \\
    V^{(l)} &= \phi(S^{(l)}) \quad \text{(Pre-FFN normalization)} \\
    F^{(l)} &= \mathrm{FF}(V^{(l)}) \quad \text{(Feed-forward)} \\
    H^{(l)} &= S^{(l)} + F^{(l)} \quad \text{(Residual connection)}
\end{align}

\subsubsection{Linear Output Network}

We are given our output dimension $d_\text{out}$. Given an a 3-layer output MLP $\theta_\text{out}$ with hidden dimension $d_\text{hidden}$, we can write our final output of the model with decoder input $D$ as:

\begin{equation}
    \mathrm{Out}(D) = \theta_\text{out}(D)
\end{equation}

\subsection{NTK Learning Scheme}

We use Neural Tangent Kernel methods to balance the loss components as in Equation \ref{eq:lossfn}. PINNs typically struggle with convergence due to the imbalanced contributions of residual, boundary, and initial losses. To address this, the model computes the NTK trace over different sets of loss terms~\citep{ntkcitation}. 

First, the training loop computes the Jacobians (gradients) of predictions with respect to model parameters for PDE residuals, initial conditions, and boundary conditions. 

For each Jacobian, we calculate the NTK trace as:

\begin{equation}
K_i = \mathrm{Tr}(J_i J_i^\top)
\end{equation}

where $J_i$ is the Jacobian matrix of the model outputs (corresponding to the $i$-th loss component) with respect to the model parameters. This measures the sensitivity of each loss component on the model's parameters during training~\citep{zhao2024pinnsformertransformerbasedframeworkphysicsinformed}. A higher $K_i$ indicates the corresponding loss component has a greater influence on the parameter updates. As a result, each loss weight is inversely proportional to the $K_i$ value. Each weight is calculated as:

\begin{equation}
    \lambda_i = \frac{\sum K}{K_i}
\end{equation}

This dynamic weighting process leads to more stable convergence during the training process. We re-compute each weight every 50 iterations in the training loop.

\section{Experiments}

\subsection{Setup}

\subsubsection{Model Ablation}
To examine the effects of the Fourier Features on the performance of the model, we created a new model called the Decoder-Only PINNsformer. This replaces the Fourier Feature embeding in the Spectral PINNsformer architecture with a single linear layer. This model serves as an ablation to examine the effects of the Fourier Features on model performance. 

\subsubsection{Data Generation}

Training and test datasets are generated via uniform sampling of collocation points over the spatial and temporal domains for each PDE. For the 1D‑reaction, convection and wave equations on \(x \in [x_\text{min}, x_\text{max}]\), \(t \in [t_\text{min}, t_\text{max}]\) as defined in \ref{pdesetup}, we generated the following data sets:

\begin{itemize}
  \item Initial condition points \(\{(x_i,0)\}_{i=1}^{N_{ic}}\) with \(x_i\) drawn uniformly from \([x_\text{min}, x_\text{max}]\).
  \item Boundary condition points \(\{(0,t_j)\}_{j=1}^{N_{bc}}\) and \(\{(2\pi,t_j)\}_{j=1}^{N_{bc}}\) with \(t_j\) drawn uniformly from \([t_\text{min},t_\text{max}]\).
  \item Residual collocation points \(\{(x_i,t_j)\}_{i=1,j=1}^{N_x,N_t}\) on a Cartesian grid of size \(N_x \times N_t\).
\end{itemize}

For the MLP‑based PINNs baseline we set \(N_{ic}=N_{bc}=101\) and \(N_x=N_t=101\). For the transformer‑based models (Pformer, S‑Pformer, DO‑Pformer) we use \(N_{ic}=N_{bc}=51\) and \(N_x=N_t=51\). Exact analytical solutions \(u_{\mathrm{true}}(x,t)\) are computed in closed form and evaluated on a test grid of \(101\times101\) points for error metrics. The network will learn the solution to the PDE by evaluating on these collocation points and enforcing the PDE via automatic differentiation. 

\paragraph{Navier–Stokes Data Generation}

We use the 2D cylinder wake dataset from \citet{RAISSI2019686}, which provides velocity fields $U_{\ast}\in\mathbb{R}^{N\times 2\times T}$ and pressure $p_{\ast}\in\mathbb{R}^{N\times T}$ at spatial locations $X_{\ast}\in\mathbb{R}^{N\times 2}$ over time $t\in\mathbb{R}^{T}$. We form a full spatio-temporal grid by combining all spatial points with all time steps, yielding $N \times T$ total points. From these, we randomly sample 2500 training points with coordinates $(x, y, t)$ as model inputs. The velocity components $(u, v)$ serve as ground truth for the loss function. For evaluation, we use the pressure field at $t=20.0$ .

Because the incompressible Navier-Stokes equations determine pressure only up to an additive constant, we align predictions with ground truth by computing the optimal offset:

\begin{equation}
    C = \frac{1}{N} \sum_{i=1}^{N}\left(p_{\text{true},i} - p_{\text{pred},i}\right)
\end{equation}

The corrected prediction $p'_{\text{pred}} = p_{\text{pred}} + C$ is used for evaluation.

\subsubsection{Network Benchmarking} \label{networkbench}
Our empirical evaluations rely on four types of PDEs: convection, 1D-reaction,
1D-wave, and Navier-Stokes, as defined in \ref{pdesetup}. For the MLP-based architecture, we uniformly sampled $N_{ic}$ = $N_{bc}$ = 101 initial and boundary points, as well as a uniform grid of $101 \times 101$ mesh points for the residual domain. In the case of training Pformer, S-Pformer and DO-Pformer, we reduce the collocation points to $N_{ic}$ = $N_{bc}$ = 51 initial and boundary points, and a uniform grid of $51 \times 51$ mesh points. For the Navier-Stokes PDE, we sample 2500 points from the residual domain for training. For all models, $d_\text{hidden}=512$, $d_\text{emb}=32$, $N=1$ (the number of decoder layers), $n_\text{heads} = 2$ (the number of attention heads). For the S-Pformer, we had a baseline of $d_\text{mapping} = 64$. The experimental setup detailed above closely matches the analysis put forth in the original PINNsformer paper for equal benchmarking~\citep{zhao2024pinnsformertransformerbasedframeworkphysicsinformed}.

\subsubsection{Evaluation}
All models were trained using the L-BFGS optimizer with Strong-Wolfe linear search for 1000 iterations. We use the L-BFGS optimizer as opposed to the more widespread Adam optimizer because of its enhanced performance in PINN optimization tasks~\citep{urbanoptimization}.

\subsubsection{Reproducibility}
All models are implemented in PyTorch~\citep{paszke2019pytorchimperativestylehighperformance}, and are trained on single NVIDIA Tesla A10G GPU.

All code is included and reproducible \href{https://github.com/rtenacity/decoder-only-pinnsformer}{at this URL}. 

\subsection{Results}

We first compare parameter counts across the PINNsformer (Pformer), Spectral PINNsformer (S-Pformer), and Decoder-Only PINNsformer (DO-Pformer) using the parameters outlined in~\ref{networkbench}.

\begin{table}[h]
  \caption{Parameter comparison between models}
  \label{param-comparison-table}
  \centering
  \begin{tabular}{ccc}
    \toprule
      PFormer   & DO-Pformer & S-Pformer      \\
    \midrule
 453,561   &    366,959  & 369,039  \\
    \bottomrule
  \end{tabular}
  \vskip -0.1in

\end{table}

The S-Pformer achieves an 18.6\% reduction in parameter count with respect to the Pformer, making it a more lightweight and efficient model by comparison.

Next, we evaluate all transformer-based models using the un-optimized parameters outlined in experimental setup across four benchmark PDEs: Convection, 1D-Reaction, 1D-Wave, and 2D Navier-Stokes. We report three metrics for each PDE: the relative Mean Absolute Error (rMAE), the relative Root Mean Squared Error (rMSE), and the training time.

\begin{table}[H]
  \caption{Comparison of transformer-based models on different PDE types, models as described in Table~\ref{param-comparison-table}}
  \label{model-comparison-table}
  \centering
  \begin{tabular}{llccc}
    \toprule
    Model     & PDE Type       & rMAE          & rMSE     & Training Time (H:MM:SS)    \\
    \midrule
    Pformer   & Convection     & 0.018         & 0.020    &  0:17:53  \\
              & 1D-Reaction    & 7.38e-3       & 0.163   & 0:03:59   \\
              & 1D-Wave        & 0.083         & 0.091     & 1:11:45   \\
              & Navier-Stokes  & 0.091         & 0.085     &  2:17:09 \\
    \midrule
    DO-Pformer& Convection     & 0.025         & 0.029    &  0:11:41  \\
              & 1D-Reaction    & 9.12e-3       & 0.020   &  0:03:40 \\
              & 1D-Wave        & 0.015         & 0.017    & 0:37:48    \\
              & Navier-Stokes  & 0.095         & 0.110 &  1:37:22  \\

    \midrule
    S-Pformer & Convection     & \textbf{0.016}   & \textbf{0.018} & 0:14:29 \\
              & 1D-Reaction    & \textbf{1.15e-3} & \textbf{2.98e-3} & 0:03:48 \\
              & 1D-Wave        & \textbf{6.94e-3} & \textbf{7.01e-3} & 0:42:40 \\
              & Navier-Stokes  & \textbf{0.079}   & \textbf{0.071} & 1:03:55 \\
    \bottomrule
  \end{tabular}
\end{table}

To further analyze the effect of Fourier features on the effectiveness of the model at handling spectral bias, we can show the error across different frequency bands of the solution of the convection PDE using a Fourier transform. We take a Fourier transform of each convection PDE solution along the spatial domain. We define the Nyquist sampling frequency $f_n$, and separate frequencies into frequency bands based on relative positions to the Nyquist frequency. We then compute the frequency error for each spectral band using mean absolute error (MAE).

\begin{table}[h]
  \caption{Error (MAE) across FFT frequency bands for different transformer-based models on convection problem, models as described in Table \ref{param-comparison-table}.}
  \label{tab:mae-frequency-bands}
  \centering
  \begin{tabular}{lccc}
    \toprule
    \multicolumn{1}{c}{} & \multicolumn{3}{c}{Error (MAE)} \\
    \cmidrule(r){2-4}
    FFT Frequency Band                             & S-Pformer & DO-Pformer & Pformer \\
    \midrule
    Very Low Frequency ($f < 0.3\,f_n$)        & 0.1401     & 0.1940     & \textbf{0.1400}   \\
    Low Frequency ($0.3\,f_n \leq f < 0.5\,f_n$)  & \textbf{0.0904}     & 0.1683     & 0.1764   \\
    Mid Frequency ($0.5\,f_n \leq f < 0.7\,f_n$)  & \textbf{0.0302}    & 0.0354     & 0.0363   \\
    High Frequency ($0.7\,f_n \leq f < 0.9\,f_n$) & \textbf{0.0110}     & 0.0157     & 0.0155   \\
    Very High Frequency ($f \geq 0.9\,f_n$)       & \textbf{0.0093}    & 0.0136     & 0.0133   \\
    \bottomrule
  \end{tabular}
\end{table}

In addition, we optimized the hyperparameters $d_\text{hidden}$, $d_\text{emb}$, $ d_\text{mapping}$ of the S-Pformer to show the full capability of the architecture. We implemented a Bayesian optimization algorithm using Optuna ~\citep{akiba2019optunanextgenerationhyperparameteroptimization} for 100 trials for each problem. The model yielding the lowest rMAE was chosen over the 100 trials. For comparison, we also optimized the $d_\text{hidden}$ and $n_\text{layers}$ of an MLP-based PINN for comparison for 100 trials using Optuna. Optimized hyperpaameters are shown in ~\ref{hyperparameteropt}.

\begin{table}[H]
  \caption{Comparison of optimized S-Pformer vs. optimized MLP-PINN performance}
  \label{tab:optimized-comparison}
  \centering
  \begin{tabular}{llcccc}
    \toprule
    Problem & Model & rMAE & rMSE & Num. Params \\
    \midrule
    Convection & MLP-PINN & 0.663 & 0.745 & \textbf{66,561}  \\
    &S-Pformer  & \textbf{0.015} & \textbf{0.018} & 305,551 \\
    \midrule
    1D-Reaction & MLP-PINN & 0.014 & 0.028 & 1,052,673  \\
    & S-Pformer  & \textbf{1.09e-3} & \textbf{2.15e-3} & \textbf{167,471} \\
    \midrule 
    1D-Wave & MLP-PINN & 0.023 & 0.023 & 2,365,441  \\
    & S-Pformer  & \textbf{2.89e-3} & \textbf{2.94e-3} & \textbf{247,823} \\
    \midrule 
    2D Navier-Stokes & MLP-PINN & \textbf{0.045} & \textbf{0.046} & 264,706  \\
    & S-Pformer  & 0.057 & 0.062 & \textbf{149,680}  \\
    \bottomrule
  \end{tabular}
\end{table}

\section{Discussion}

Our results demonstrate that architectural simplification yields superior performance in physics-informed neural networks. The S-Pformer achieves consistent improvements compared to transformer-based architectures across all benchmark PDEs while reducing parameter count by 18.6\%, challenging the conventional "bigger is better" paradigm in physics-based deep learning. In addition, the optimized S-Pformer shows improved or comparable performance compared to an optimized MLP-based PINN, while using a fraction of the parameter count. 

The frequency band analysis (Table~\ref{tab:mae-frequency-bands}) provides direct evidence that Fourier features effectively address spectral bias - a persistent limitation of both traditional PINNs and the original PINNsformer. The 30\% error reduction in high-frequency regimes $(f > 0.7f_n)$ compared to the decoder-only baseline confirms that explicit frequency encoding is important for capturing multiscale PDE behaviors. This improvement is  significant for problems like the convection equation where high-frequency dynamics dominate the solution.

The decoder-only design proves that the encoder in the original PINNsformer introduced unnecessary computational overhead without corresponding performance gains. By applying self-attention directly to embedded coordinates, we maintain the temporal dependency modeling that makes transformers effective for time-dependent PDEs while eliminating redundant computation. This efficiency gain becomes more pronounced with increasing problem complexity, as evidenced by consistently shorter training times.

The S-Pformer demonstrates consistent versatility across elliptic, parabolic, and hyperbolic PDEs. This robustness stems from the adaptive nature of the Fourier feature mapping, which learns optimal frequency representations for each problem type rather than relying on fixed spectral assumptions. 

While our optimized S-Pformer variants (Table ~\ref{tab:optimized-s-pformer}) show the architecture's full potential, the hyperparameter sensitivity suggests room for more principled hyperparameter choices. While we optimized key architectural parameters ($d_\text{hidden}$, $d_\text{emb}$, $d_\text{mapping}$), future work should explore the sensitivity to attention head count, though preliminary experiments suggest $n_\text{heads}=2$ provides a good efficiency-performance trade-off.

The Navier-Stokes results reveal a constraint: while the S-Pformer excels on physics-informed problems with purely automatic-differentiation-based losses, it shows marginal underperformance compared to MLPs on data-driven components such as the Navier-Stokes benchmark. 

Future work should investigate adaptive frequency selection mechanisms, extend evaluation to more complex geometries and coupled systems, and explore approaches that combine the strengths of both transformer and MLP architectures for different physics-based and data-driven components of the PINN loss function.

\bibliographystyle{plainnat}  
\bibliography{references}     

\newpage

\appendix

\section{Appendix}

\subsection{PINN Architecture Computational Performance Comparison}
We took every model type and computed metrics to evaluate computational performance, in this case on the 1D-Reaction PDE. 

\begin{table}[H]
  \caption{Computational Performance Comparison of PINN Architectures}
  \label{tab:pinn-comparison}
  \centering
  \begin{tabular}{lcccc}
    \toprule
    Model & Avg. Step Time (s) & Avg. GPU Mem. (MB) & Params & MFLOPs \\
    \midrule
    MLP-based PINN &  0.39 & 430.5 & 527,361 & 5.28 \\
    Pformer        & 0.54  & 457.7 & 453,561 & 4.54 \\
    DO-Pformer     & 0.48 & 363.5 & 366,959 & 3.68 \\
    S-Pformer      & 0.50 & 372.5 & 370,991 & 3.72 \\
    \bottomrule
  \end{tabular}
\end{table}

\subsection{Hyperparameter Optimization} \label{hyperparameteropt}

We optimized the hyperparameters $d_\text{hidden}$, $d_\text{emb}$, $ d_\text{mapping}$ of the S-Pformer in Optuna for 100 trials for each problem. The model yielding the lowest rMAE was chosen over the 100 trials. 

\begin{table}[h]
  \caption{Optimized S-Pformer Performance}
  \label{tab:optimized-s-pformer}
  \centering
  \begin{tabular}{lcccccc}
    \toprule
    Problem & rMAE & rMSE & $d_{\text{hidden}}$ & $d_{\text{emb}}$ & $d_{\text{mapping}}$  & Num. Params \\
    \midrule
    Convection & 0.015 & 0.018 & 256 & 128 & 32 & 305,551 \\
    1D-Reaction  & 1.09e-3 & 2.15e-3 & 256 & 32 & 96 & 167,471 \\
    1D-Wave    & 2.89e-3 & 2.94e-3 & 128 & 128 & 64 & 247,823 \\
    2D Navier-Stokes   & 0.057 & 0.062 & 256 & 16 & 112 & 149,680 \\
    \bottomrule
  \end{tabular}
\end{table}

We also optimized the $d_\text{hidden}$ and $n_\text{layers}$ of an MLP-based PINN for comparison for 100 trials using Optuna. 

\begin{table}[H]
  \caption{Optimized MLP-PINN Performance}
  \label{tab:optimized-pinn}
  \centering
  \begin{tabular}{lcccccc}
    \toprule
    Problem & rMAE & rMSE & $d_{\text{hidden}}$ & $n_\text{layers}$ & Num. Params \\
    \midrule
    Convection & 0.663 & 0.745 & 128 & 6 & 66,561  \\
    1D-Reaction  & 0.014 & 0.028 & 512 & 6 & 1,052,673 \\
    1D-Wave    & 0.023 & 0.023 & 768 & 6 & 2,365,441 \\
    2D Navier-Stokes   & 0.045 & 0.046 & 256 & 6 & 264,706 \\ 
    \bottomrule
  \end{tabular}
\end{table}

\subsection{PDE Equations}\label{pdesetup}

\paragraph{Convection} The one-dimensional convection problem is a hyperbolic PDE used to model transfer processes.

\begin{equation}
    \frac{\partial u}{\partial t} + \beta \frac{\partial u}{\partial x} = 0, 
    \quad \forall x \in [0, 2\pi], \; t \in [0, 1]
\end{equation}

\begin{equation}
    \text{IC:} \quad u(x, 0) = \sin(x) \qquad 
    \text{BC:} \quad u(0, t) = u(2\pi, t)
\end{equation}

where $\beta$ is the convection coefficient. In this case, $\beta=50$. This is a high-frequency PDE, which makes it difficult for conventional PINNs to approximate. 

\paragraph{1D-Reaction}
The one-dimensional reaction problem is a hyperbolic PDE used to model chemical reactions.

\begin{equation}
    \frac{\partial u}{\partial t} - \rho u(1 - u) = 0, \quad \forall x \in [0, 2\pi], \; t \in [0, 1]
\end{equation}

\begin{equation}
    \text{IC:} \quad u(x, 0) = \exp\left(-\frac{(x - \pi)^2}{2(\pi/4)^2}\right), \qquad 
    \text{BC:} \quad u(0, t) = u(2\pi, t)
\end{equation}

where $\rho$ is the reaction coefficient, where $\rho = 5$. The equation has an analytical solution:

\begin{equation}
    u_{\text{analytical}} = \frac{h(x)\exp(\rho t)}{h(x)\exp(\rho t) + 1 - h(x)}
\end{equation}

where $h(x)$ is the function of the initial condition.

\paragraph{1D-Wave PDE}
The 1D-Wave equation is a hyperbolic PDE that is used to describe the propagation of waves in one spatial dimension.

\begin{equation}
    \frac{\partial^2 u}{\partial t^2} - \beta \frac{\partial^2 u}{\partial x^2} = 0 \quad \forall x \in [0, 1], \; t \in [0, 1]
\end{equation}
\begin{equation}
    \text{IC:} \quad u(x, 0) = \sin(\pi x) + \frac{1}{2} \sin(\beta \pi x), \quad \frac{\partial u(x, 0)}{\partial t} = 0
\end{equation}

\begin{equation}
    \text{BC:} \quad u(0, t) = u(1, t) = 0
\end{equation}

where $\beta$ is the wave speed, where $\beta = 3$. The equation has an analytical solution:

\begin{equation}
    u(x, t) = \sin(\pi x)\cos(2\pi t) + \frac{1}{2} \sin(3\pi x) \cos(6\pi t)
\end{equation}

\paragraph{2D Navier-Stokes PDE.} The 2D Navier-Stokes equation is a parabolic PDE that consists of a pair of partial differential equations that describe the behavior of incompressible fluid flow in two-dimensional space.
\[
\begin{aligned}
    \frac{\partial u}{\partial t} + \lambda_1 \left( u \frac{\partial u}{\partial x} + v \frac{\partial u}{\partial y} \right) &= -\frac{\partial p}{\partial x} + \lambda_2 \left( \frac{\partial^2 u}{\partial x^2} + \frac{\partial^2 u}{\partial y^2} \right) \\
    \frac{\partial v}{\partial t} + \lambda_1 \left( u \frac{\partial v}{\partial x} + v \frac{\partial v}{\partial y} \right) &= -\frac{\partial p}{\partial y} + \lambda_2 \left( \frac{\partial^2 v}{\partial x^2} + \frac{\partial^2 v}{\partial y^2} \right)
\end{aligned}
\]

where $u(t,x,y)$ and $v(t,x,y)$ are the $x$-component and $y$-component of the velocity field separately, and $p(t,x,y)$ is the pressure. Here, $\lambda_1 = 1$ and $\lambda_2 = 0.01$.  The simulated solution is given by~\citep{RAISSI2019686}.

\newpage

\subsection{Visualization of Spectral PINNsformer}

\begin{figure}[h]
    \centering

    \begin{subfigure}[b]{0.3\textwidth}
        \includegraphics[width=\linewidth]{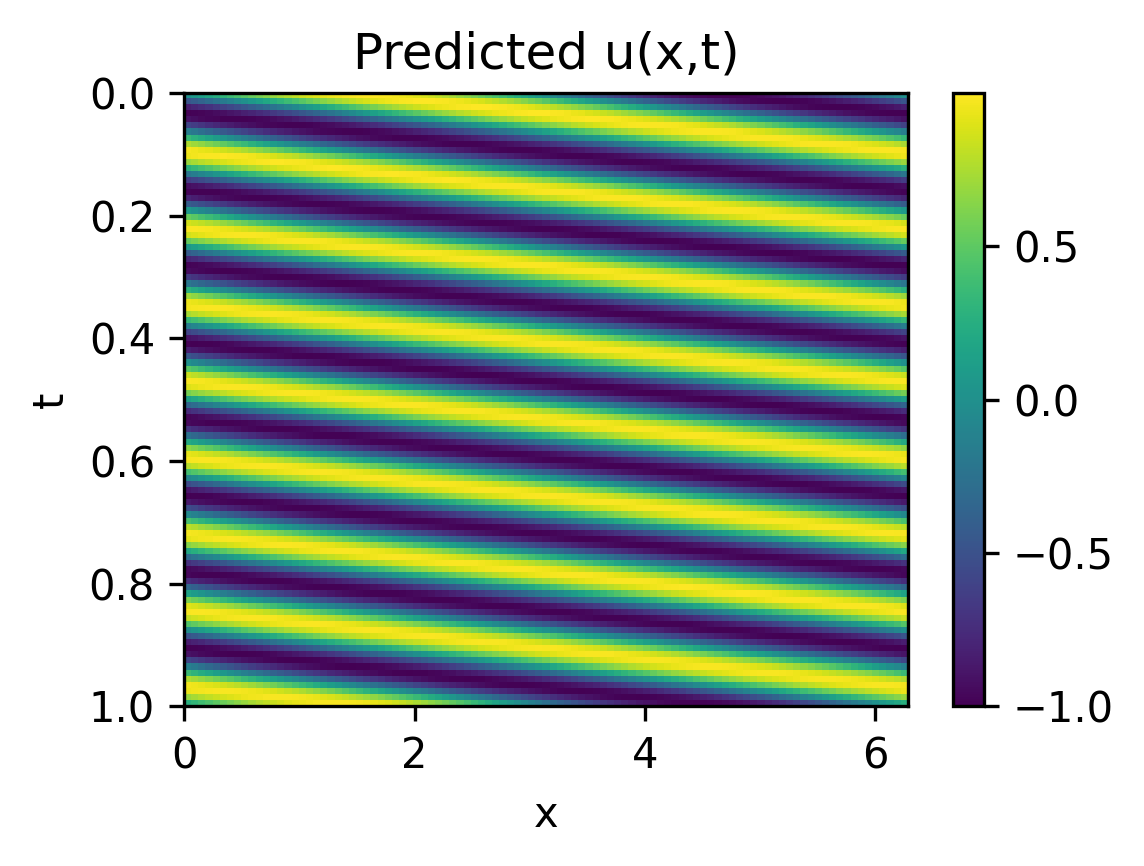}
        \caption{Pred - Convection}
    \end{subfigure}
    \begin{subfigure}[b]{0.3\textwidth}
        \includegraphics[width=\linewidth]{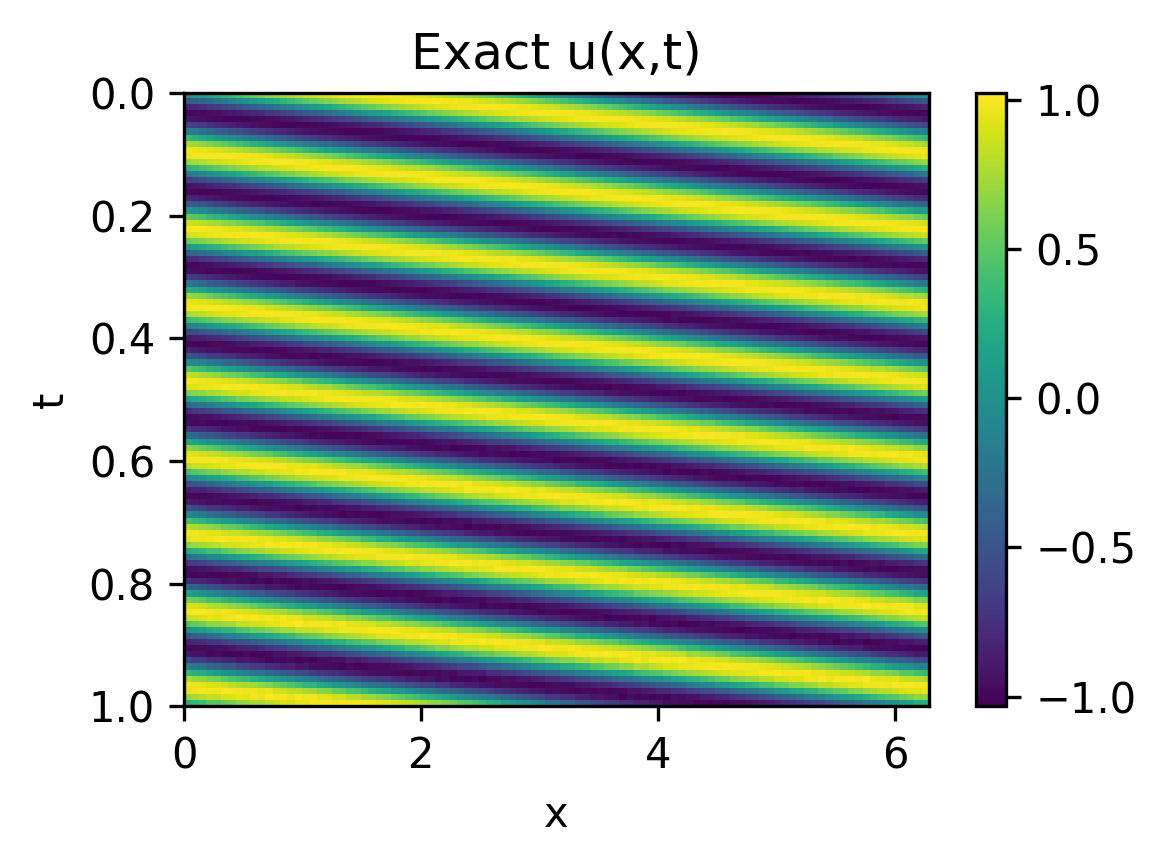}
        \caption{Exact - Convection}
    \end{subfigure}
    \begin{subfigure}[b]{0.3\textwidth}
        \includegraphics[width=\linewidth]{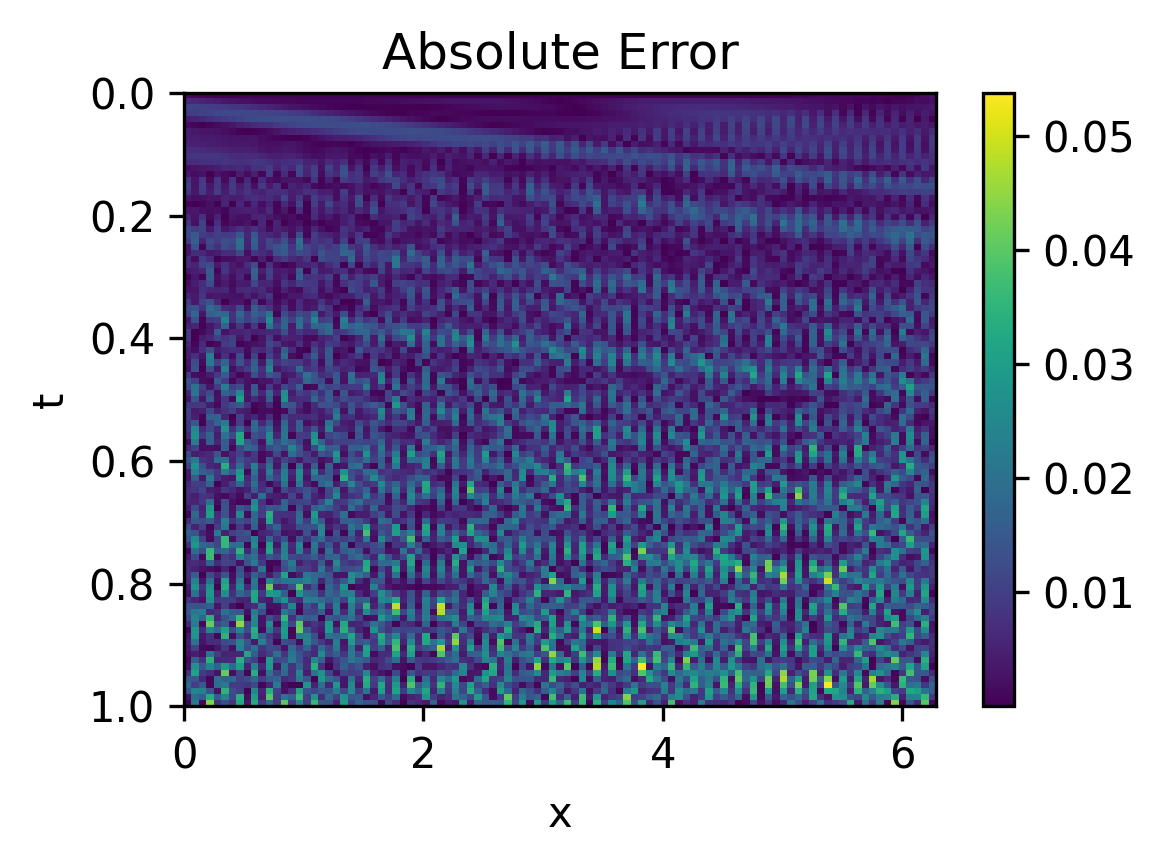}
        \caption{Error - Convection}
    \end{subfigure}

    \hspace{1em}

    \begin{subfigure}[b]{0.3\textwidth}
        \includegraphics[width=\linewidth]{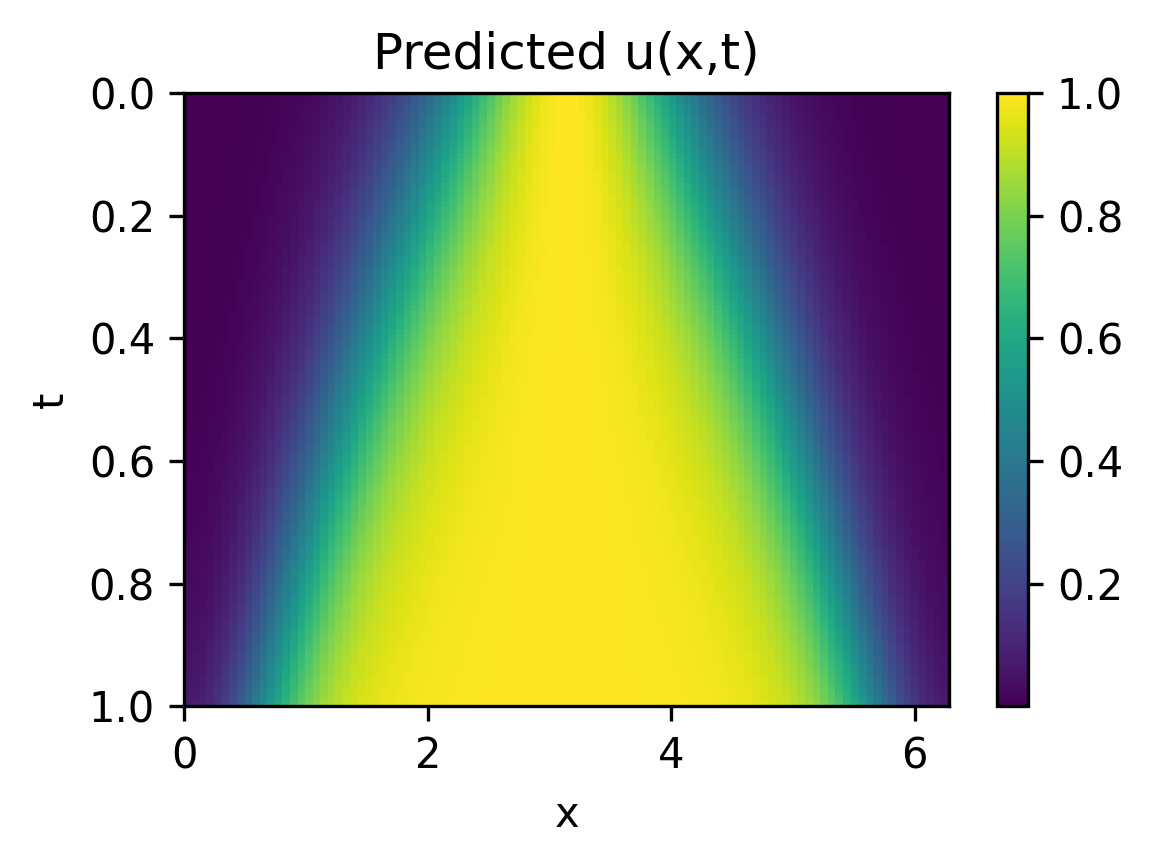}
        \caption{Pred - 1D Reaction}
    \end{subfigure}
    \begin{subfigure}[b]{0.3\textwidth}
        \includegraphics[width=\linewidth]{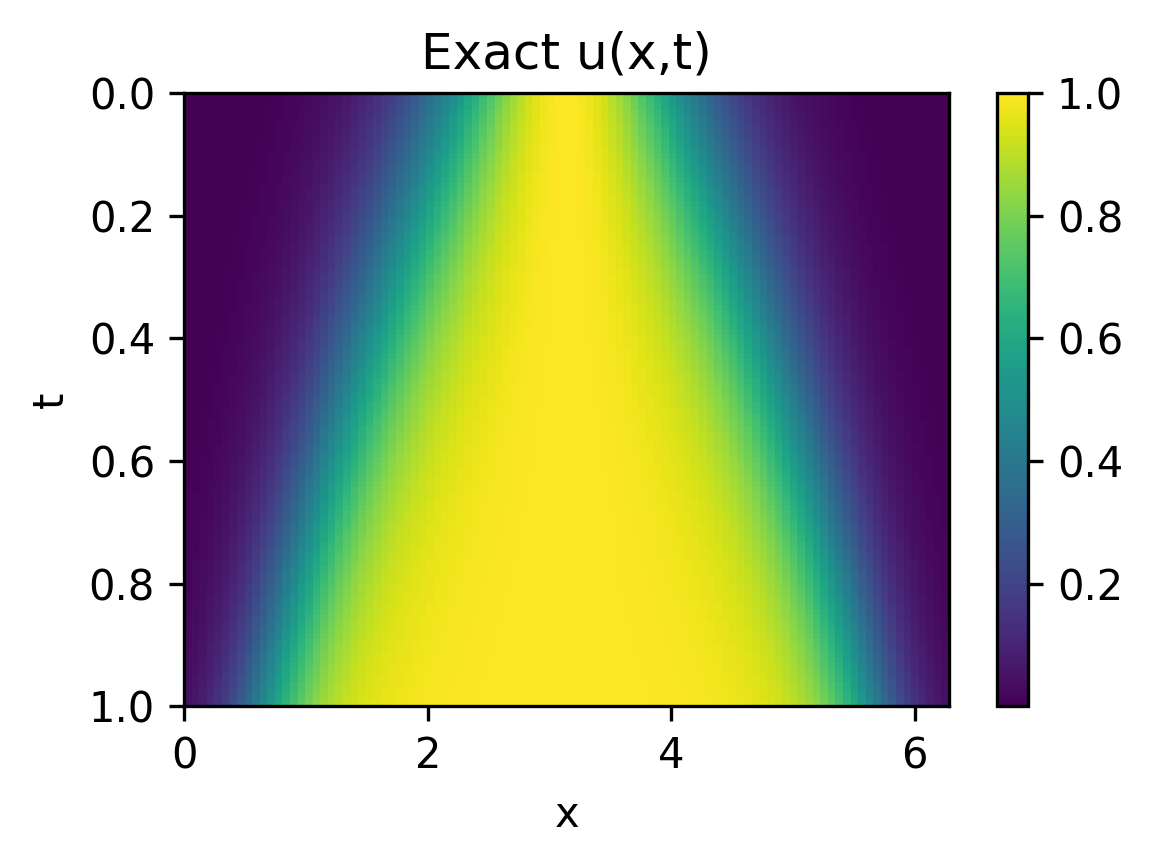}
        \caption{Exact - 1D Reaction}
    \end{subfigure}
    \begin{subfigure}[b]{0.3\textwidth}
        \includegraphics[width=\linewidth]{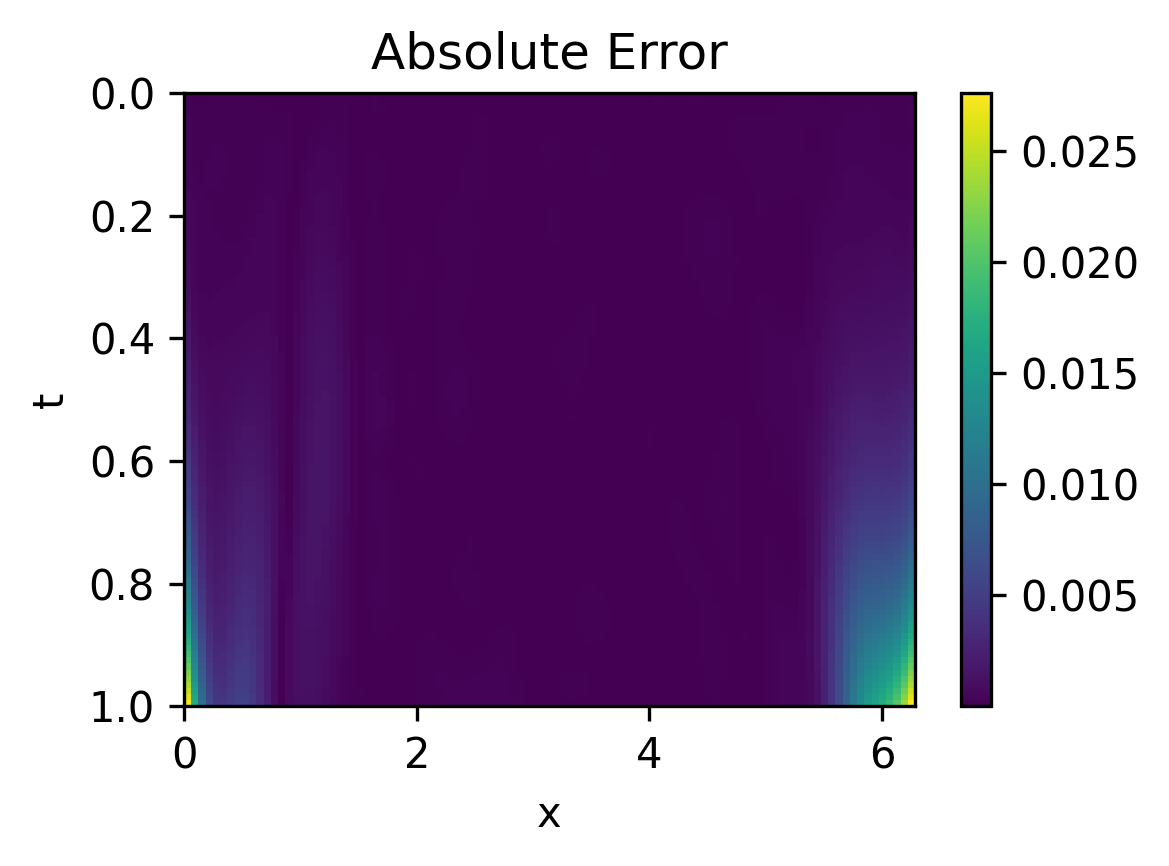}
        \caption{Error - 1D Reaction}
    \end{subfigure}

    \hspace{1em}

    \begin{subfigure}[b]{0.3\textwidth}
        \includegraphics[width=\linewidth]{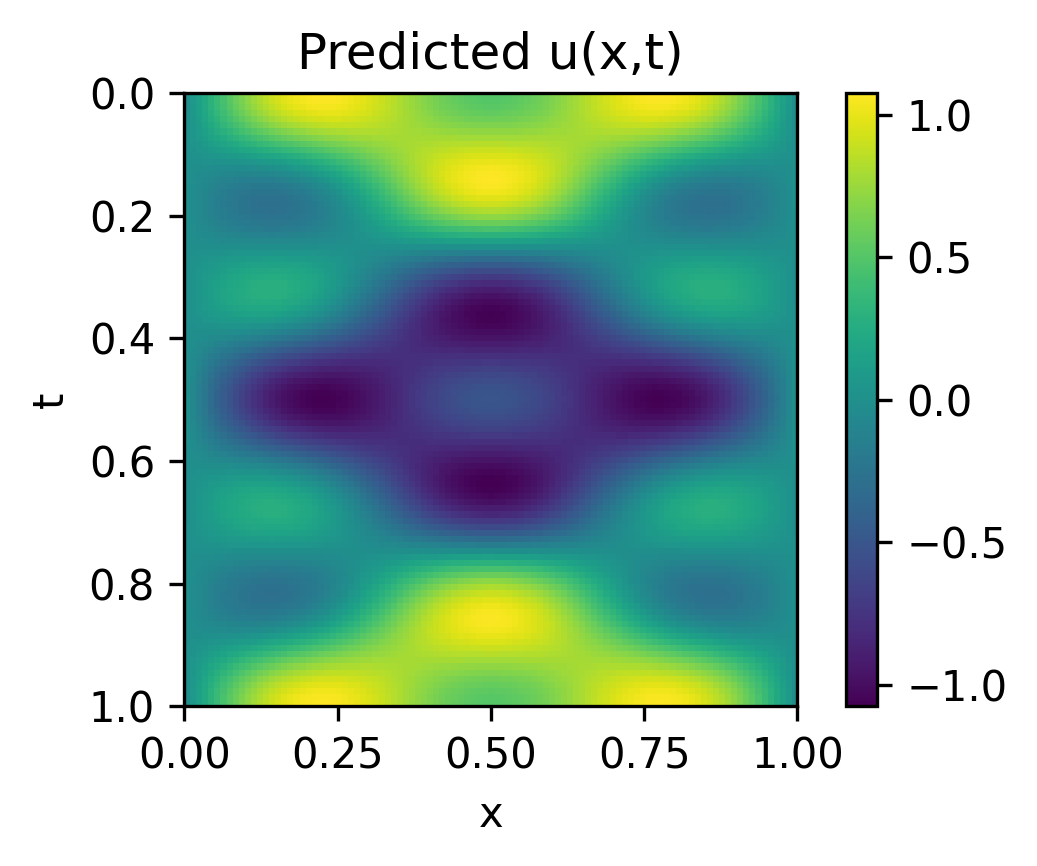}
        \caption{Pred - 1D Wave}
    \end{subfigure}
    \begin{subfigure}[b]{0.3\textwidth}
        \includegraphics[width=\linewidth]{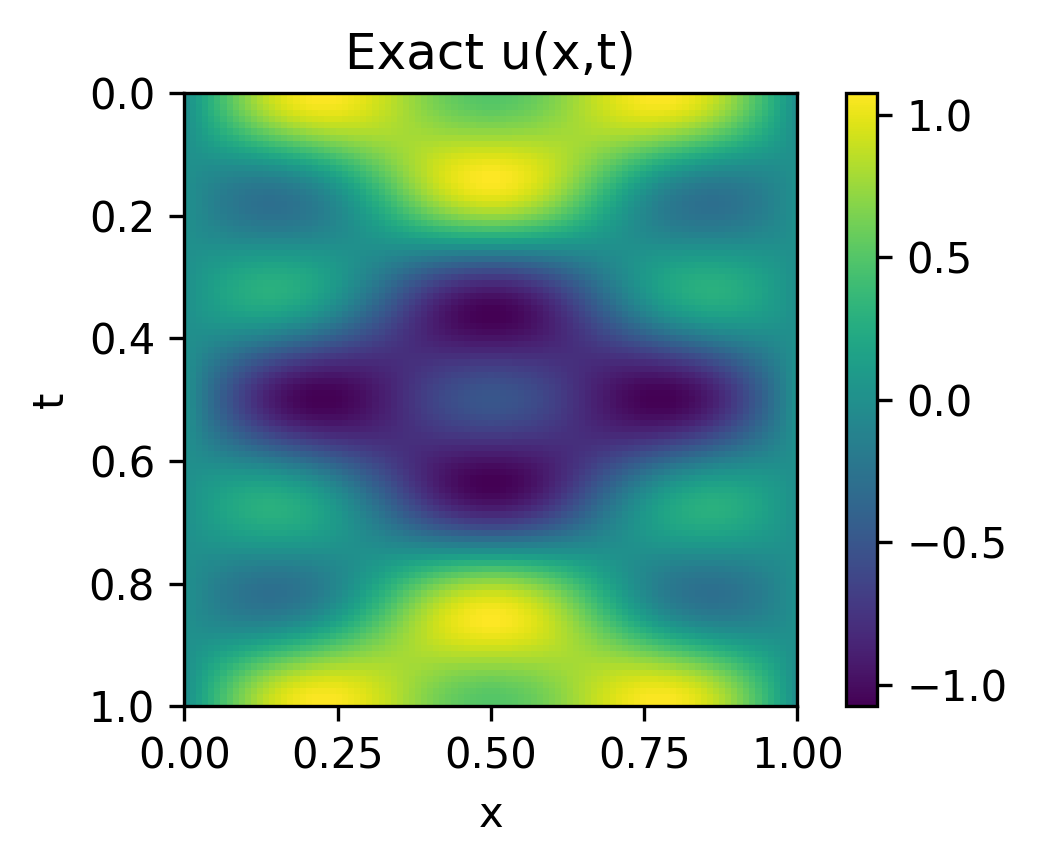}
        \caption{Exact - 1D Wave}
    \end{subfigure}
    \begin{subfigure}[b]{0.3\textwidth}
        \includegraphics[width=\linewidth]{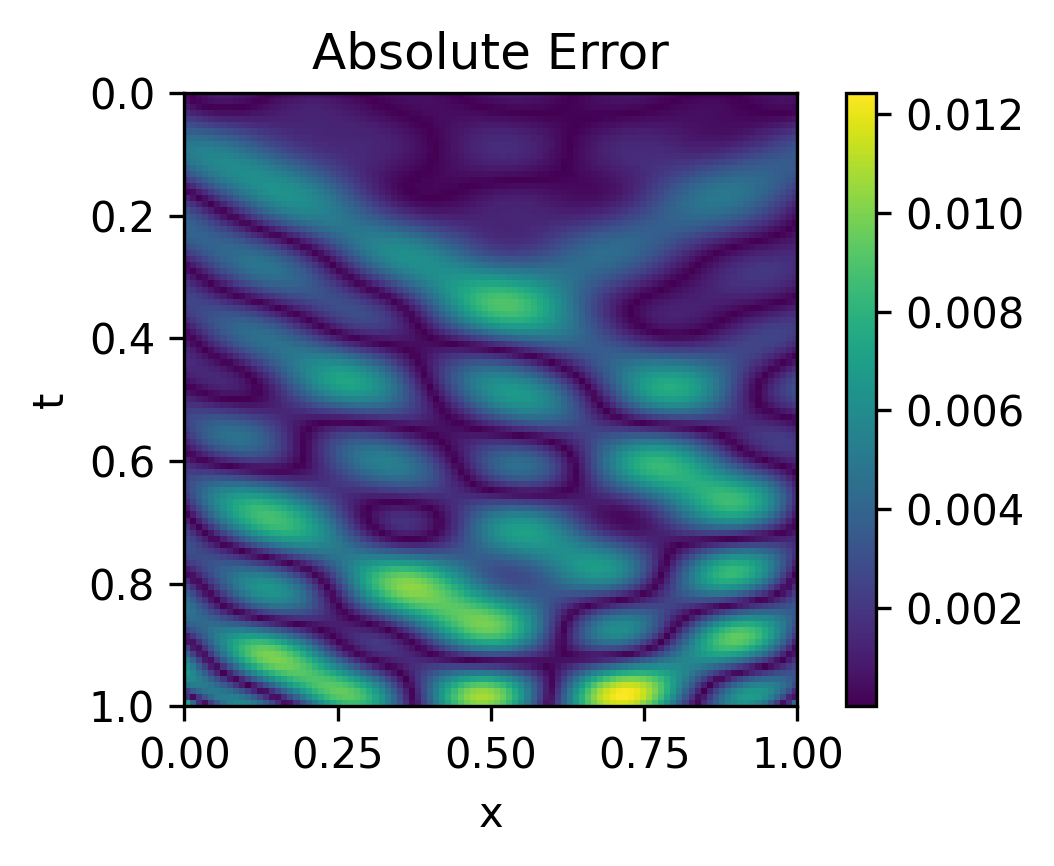}
        \caption{Error - 1D Wave}
    \end{subfigure}

    \hspace{1em}

    \begin{subfigure}[b]{0.3\textwidth}
        \includegraphics[width=\linewidth]{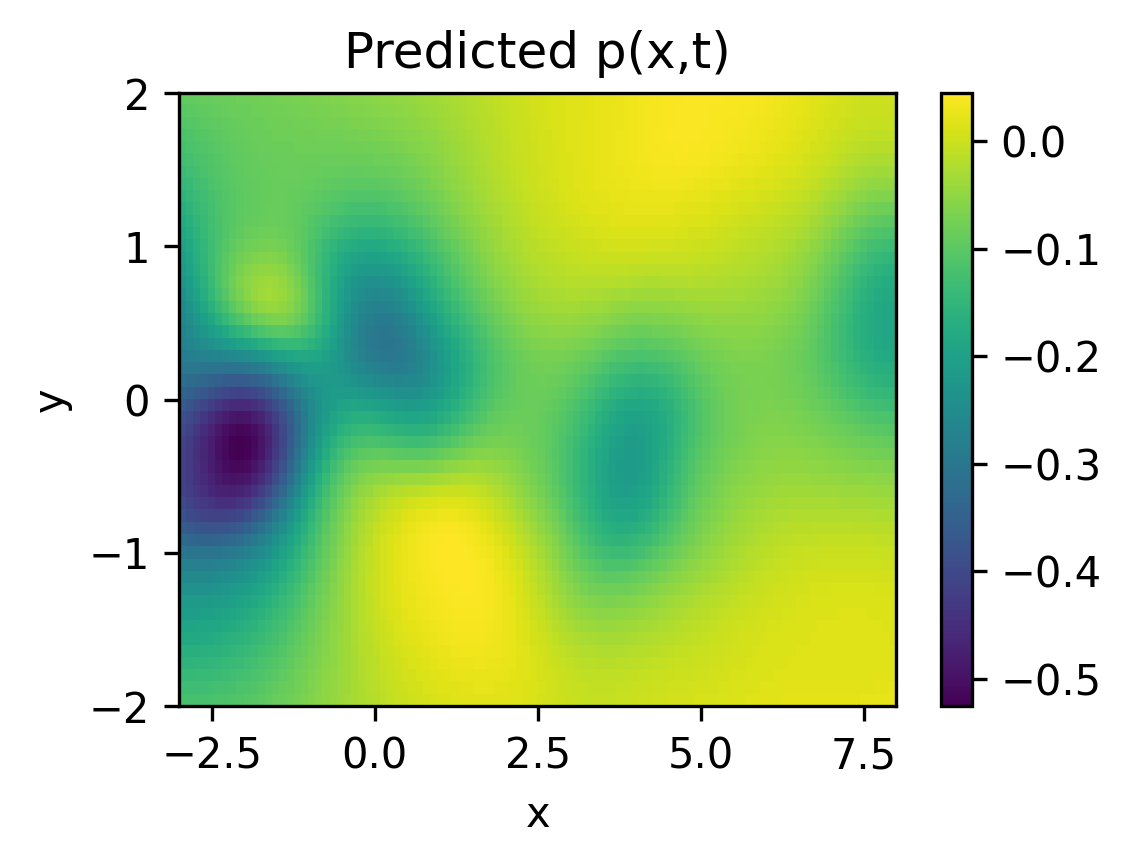}
        \caption{Pred - 2D N-S ($t=20$)}
    \end{subfigure}
    \begin{subfigure}[b]{0.3\textwidth}
        \includegraphics[width=\linewidth]{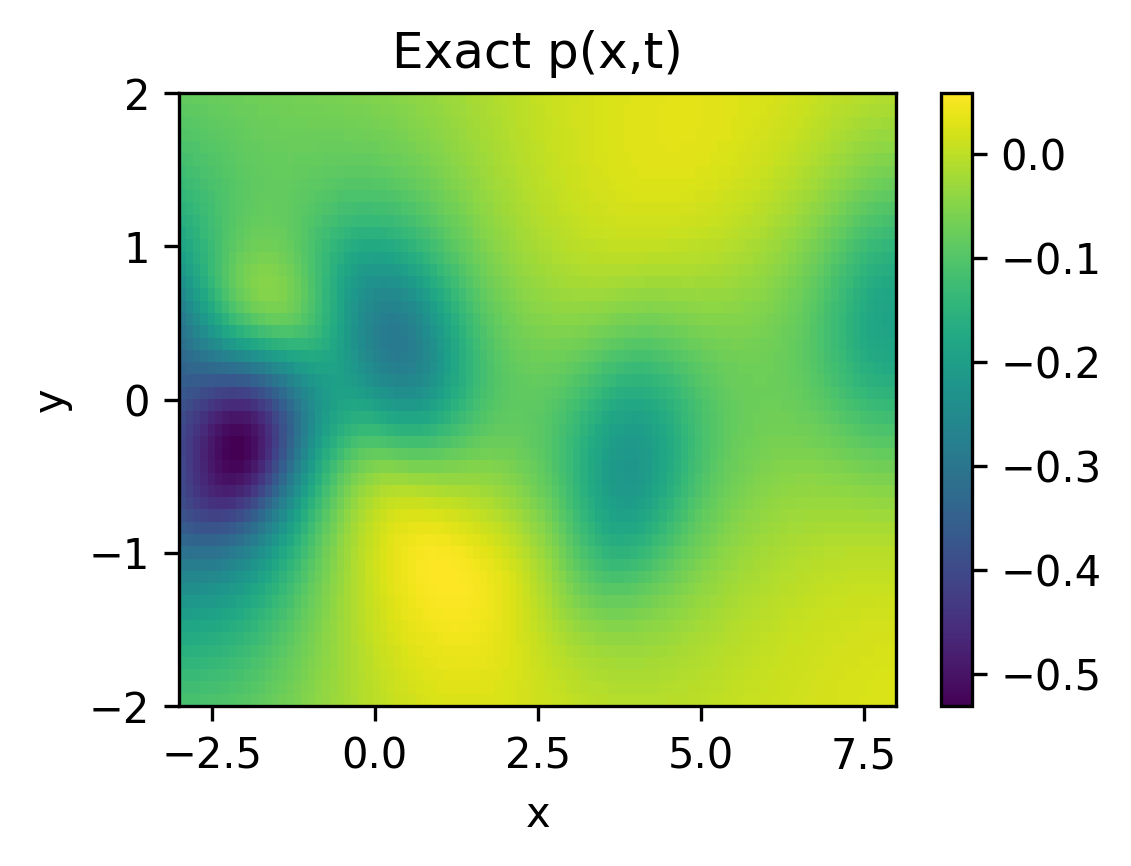}
        \caption{Exact - 2D N-S ($t=20$) }
    \end{subfigure}
    \begin{subfigure}[b]{0.3\textwidth}
        \includegraphics[width=\linewidth]{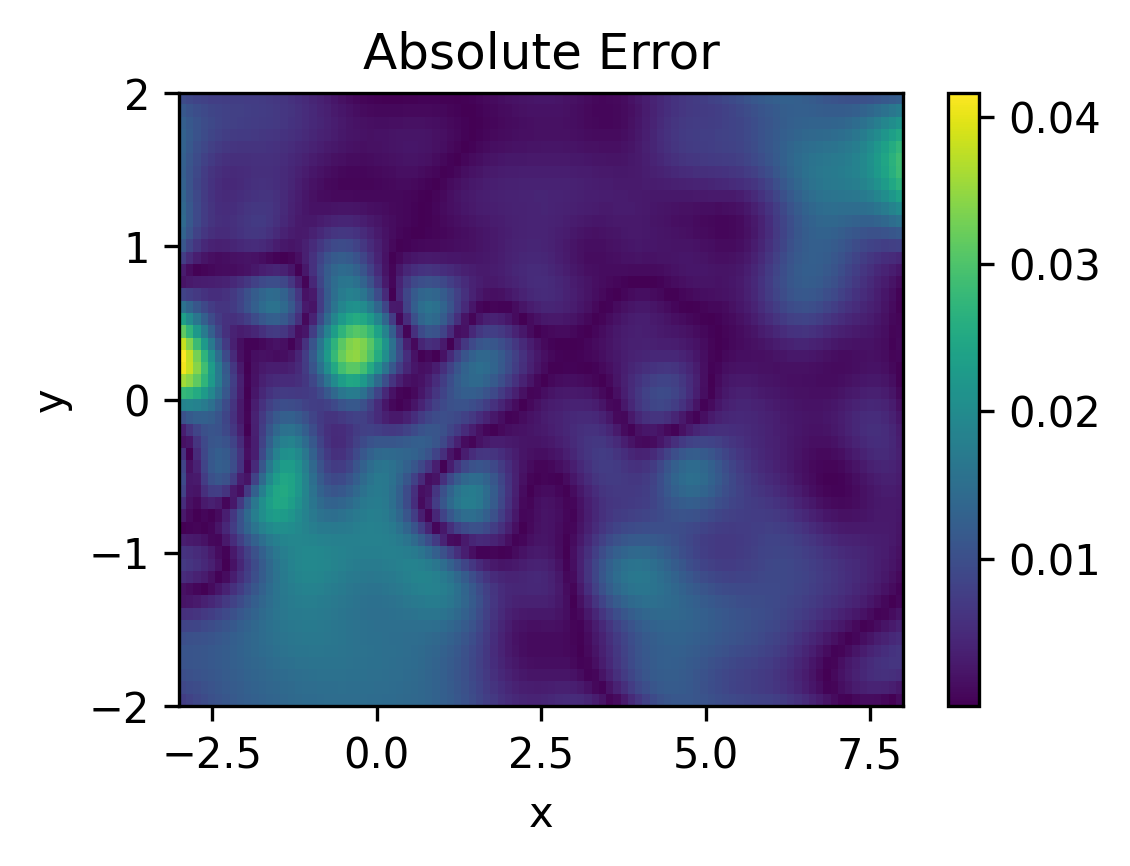}
        \caption{Error - 2D N-S ($t=20$)}
    \end{subfigure}

    \caption{
    In the figure above, the first column shows the S-Pformer prediction, the middle column shows the ground truth, and the last column shows the prediction error.}
\end{figure}

\newpage

\subsection{Loss Convergence}

To examine the stability and convergence of the loss as all models trained, we plotted the loss for each optimizer step during the training process. 

\subsubsection{Convection Loss}

\begin{figure}[h]
    \centering
    \begin{subfigure}[b]{0.45\textwidth}
        \includegraphics[width=\linewidth]{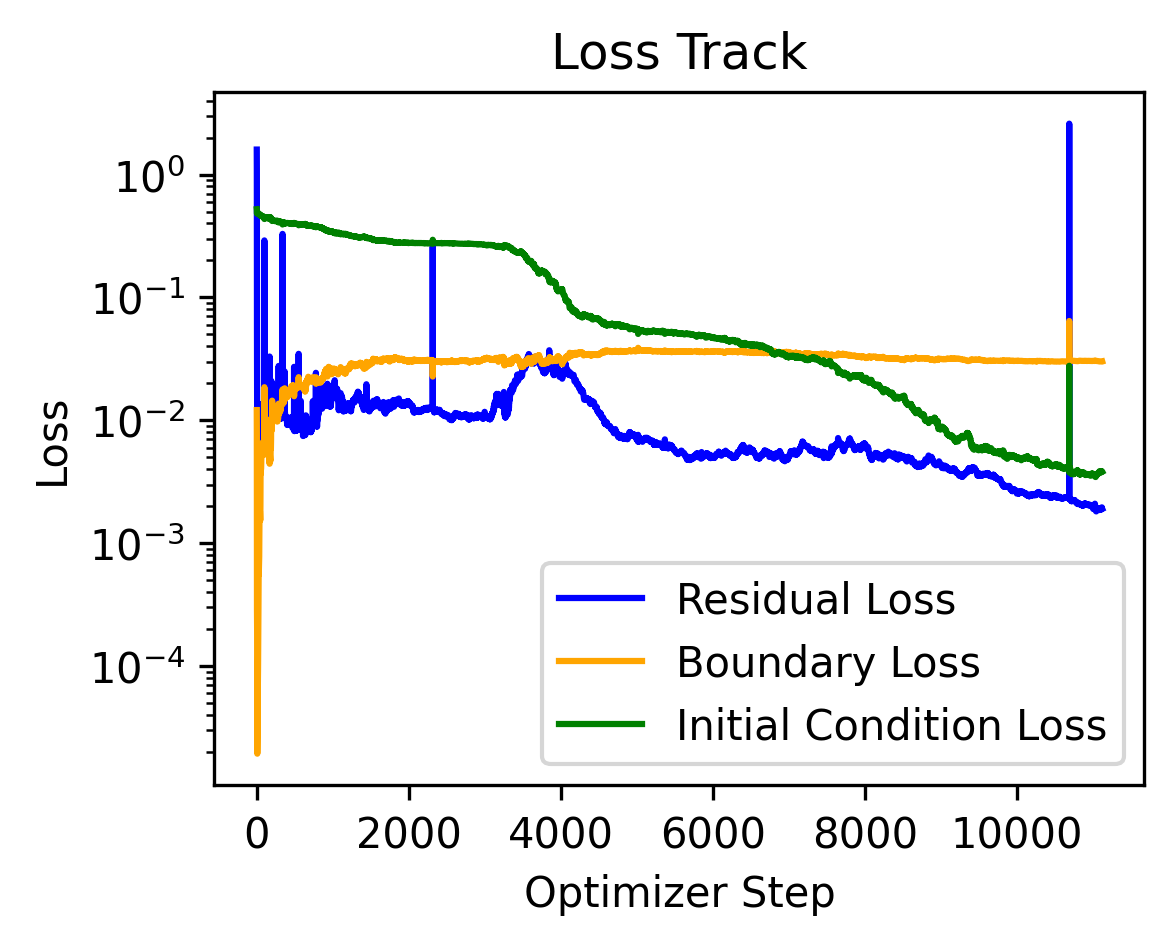}
        \caption{Loss - PINN}
    \end{subfigure}
    \begin{subfigure}[b]{0.45\textwidth}
        \includegraphics[width=\linewidth]{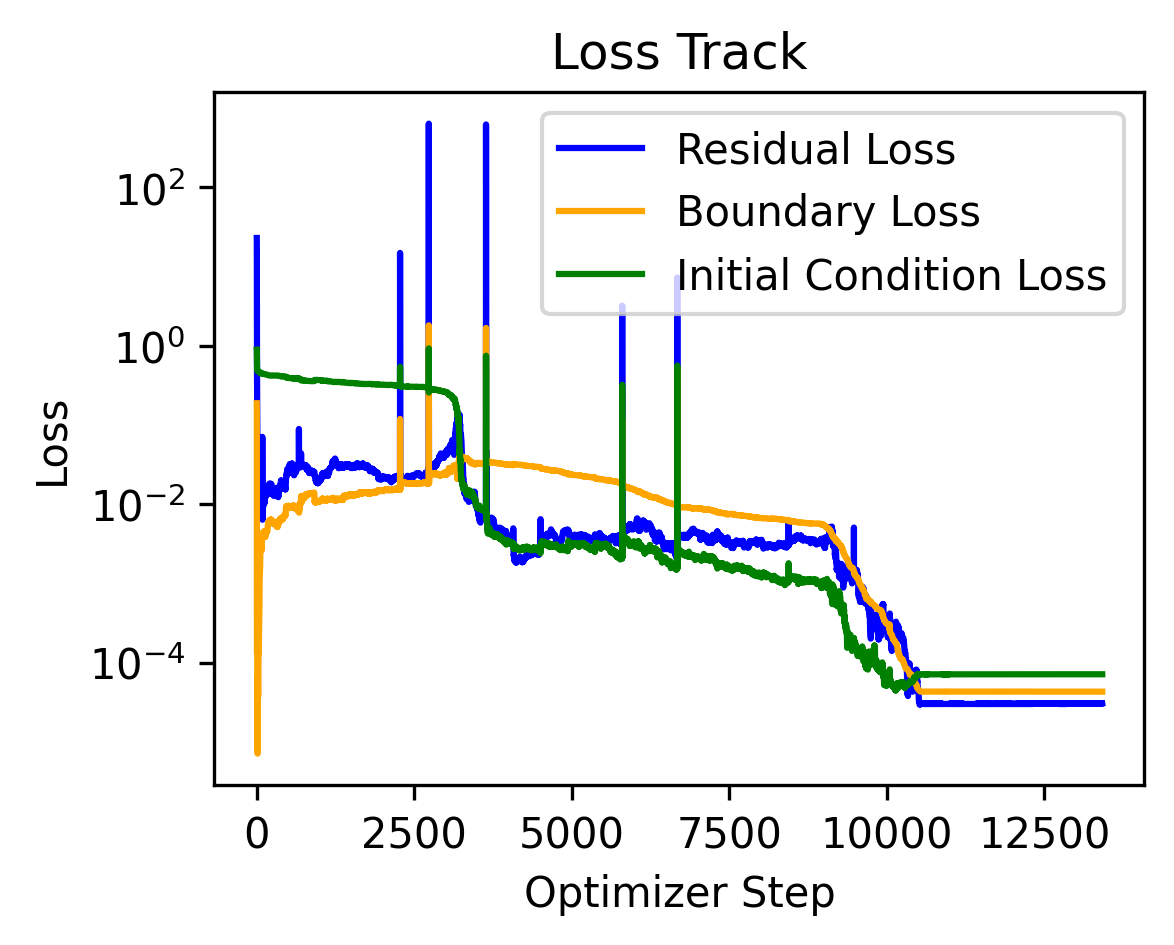}
        \caption{Loss - Pformer}
    \end{subfigure}

\end{figure}

\begin{figure}[h]
    \centering
    \begin{subfigure}[b]{0.45\textwidth}
        \includegraphics[width=\linewidth]{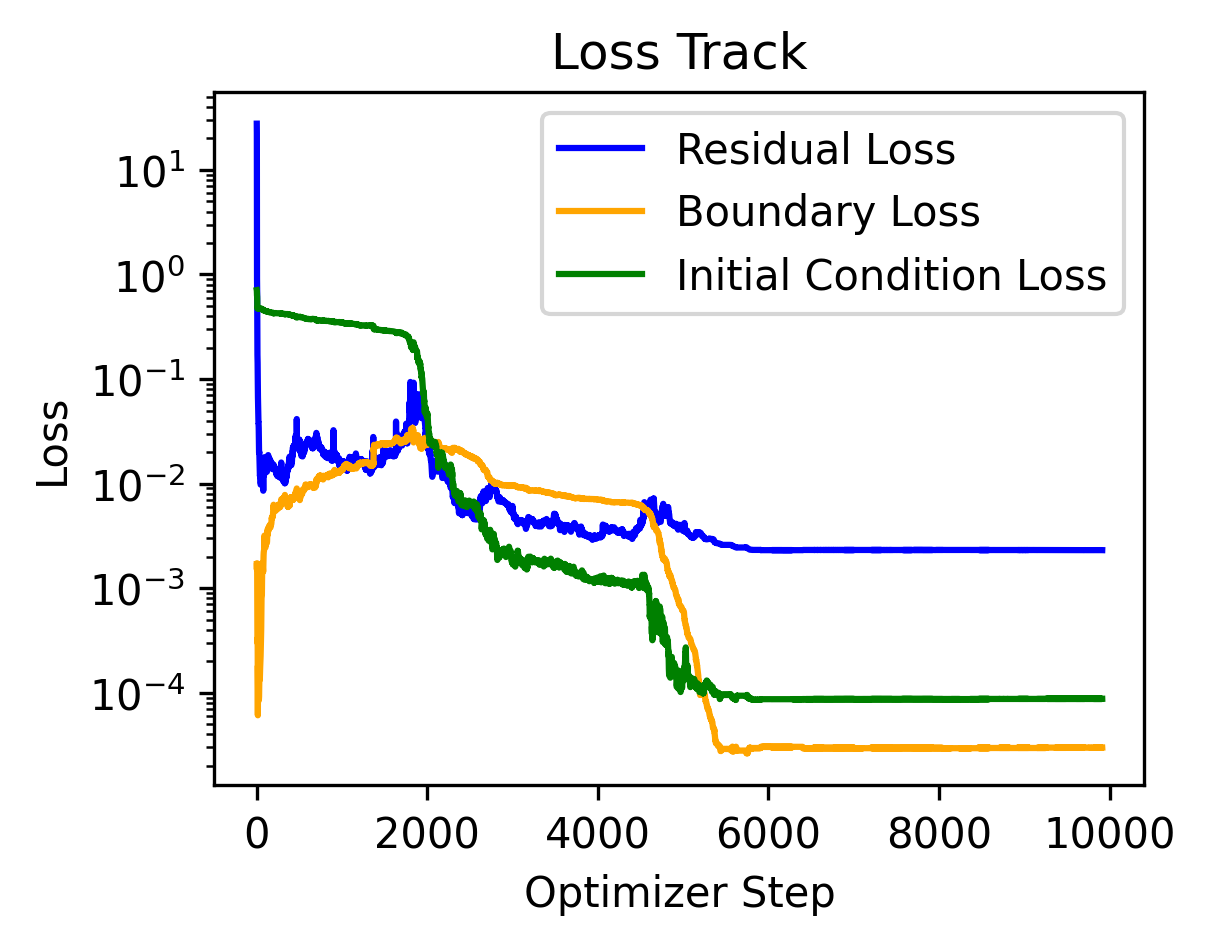}
        \caption{Loss - DO-Pformer}
    \end{subfigure}
    \begin{subfigure}[b]{0.45\textwidth}
        \includegraphics[width=\linewidth]{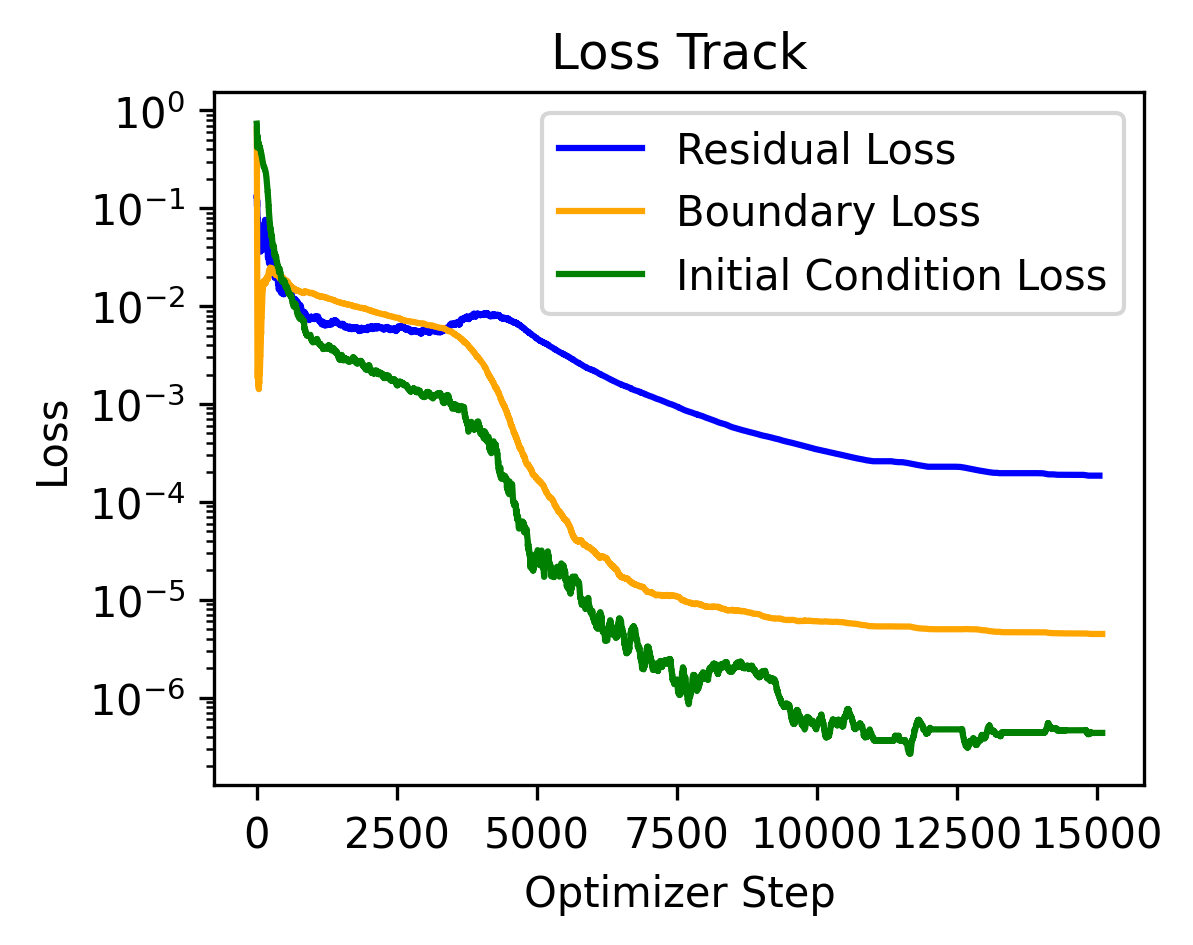}
        \caption{Loss - S-Pformer}
    \end{subfigure}

\end{figure}

\newpage

\subsubsection{1D Reaction Loss}

\begin{figure}[H]
    \centering
    \begin{subfigure}[b]{0.45\textwidth}
        \includegraphics[width=\linewidth]{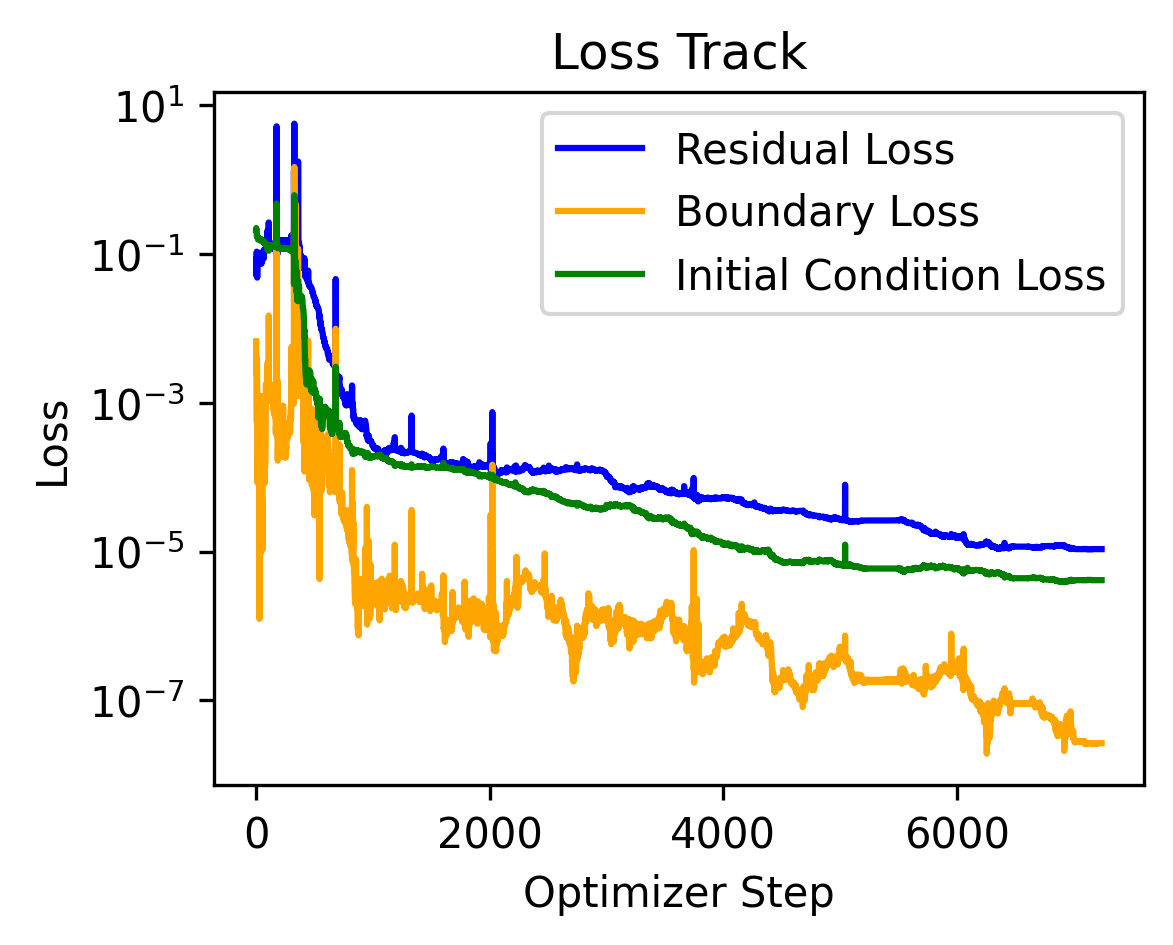}
        \caption{Loss - PINN}
    \end{subfigure}
    \begin{subfigure}[b]{0.45\textwidth}
        \includegraphics[width=\linewidth]{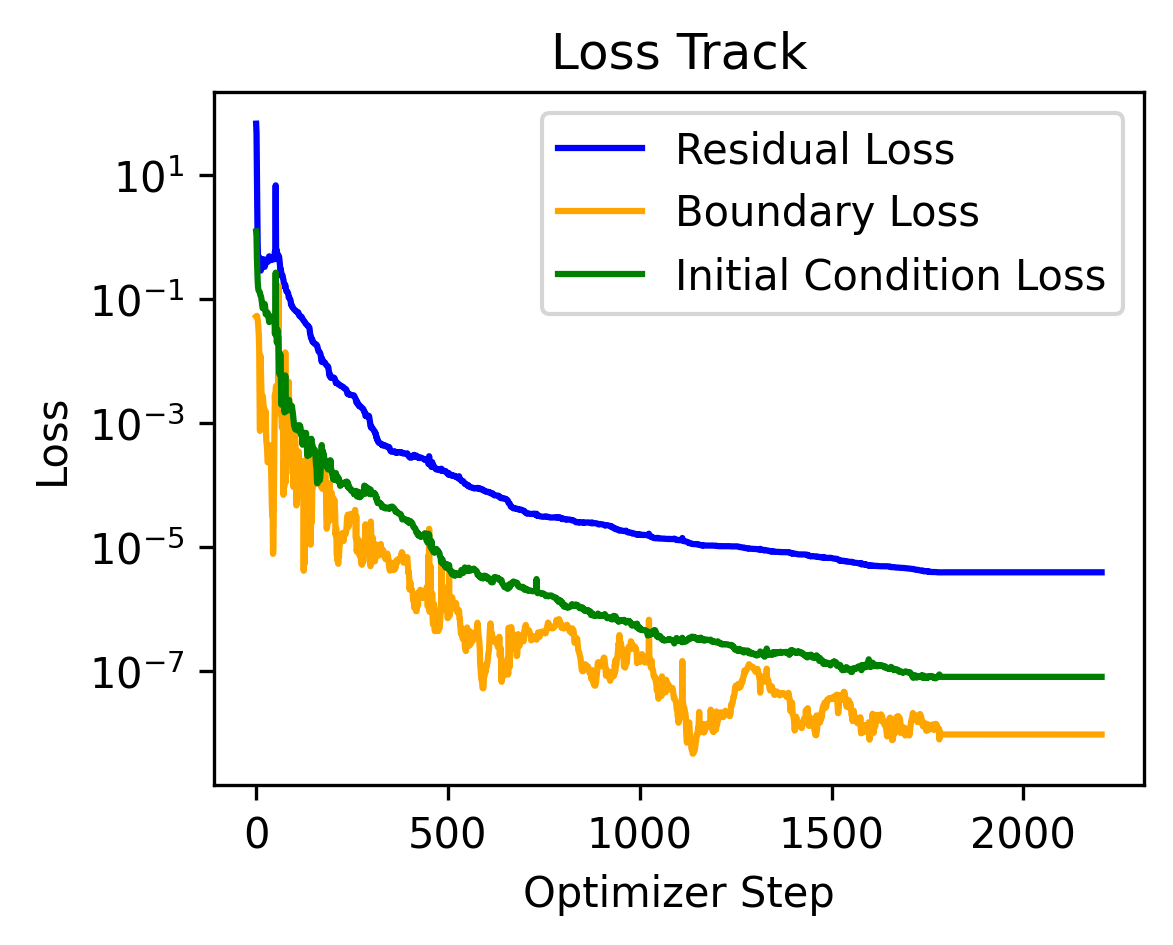}
        \caption{Loss - Pformer}
    \end{subfigure}

\end{figure}

\begin{figure}[H]
    \centering
    \begin{subfigure}[b]{0.45\textwidth}
        \includegraphics[width=\linewidth]{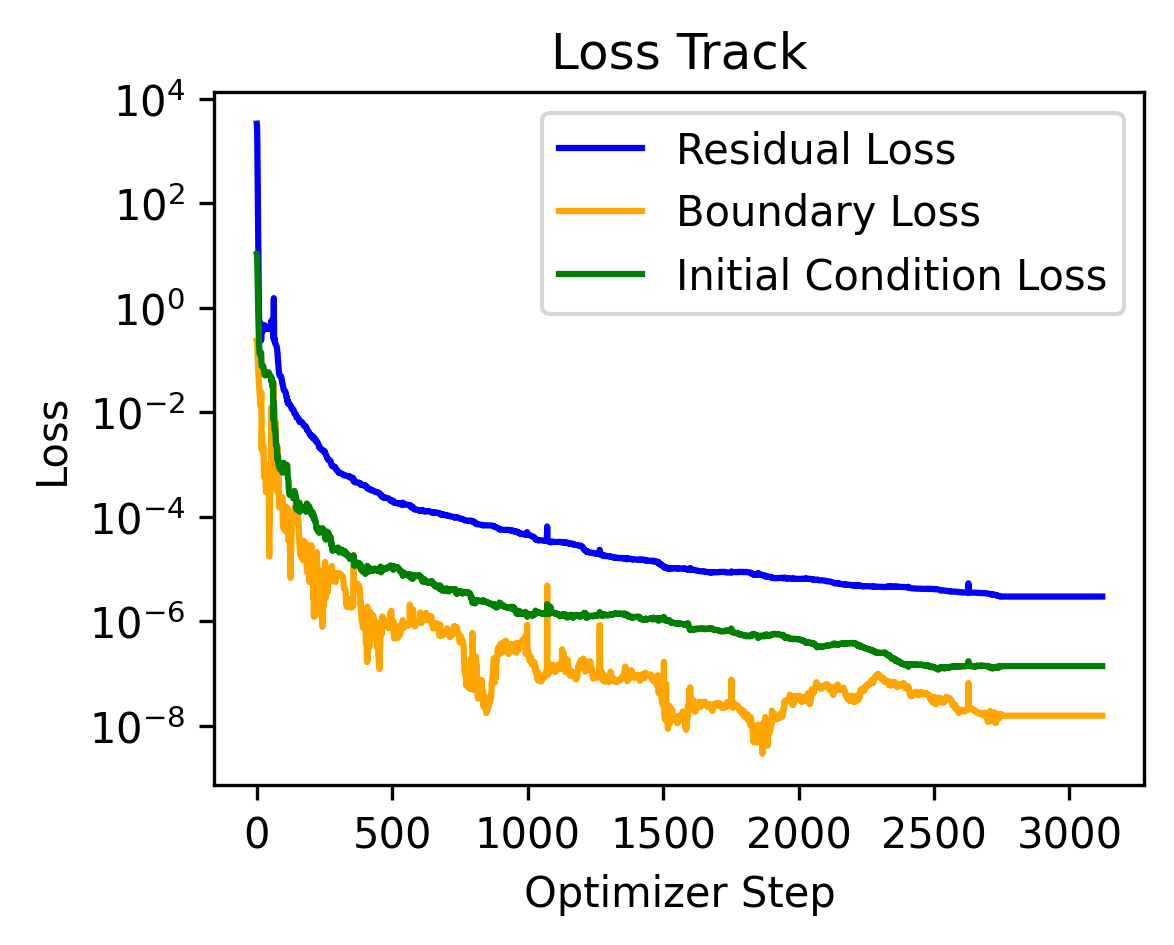}
        \caption{Loss - DO-Pformer}
    \end{subfigure}
    \begin{subfigure}[b]{0.45\textwidth}
        \includegraphics[width=\linewidth]{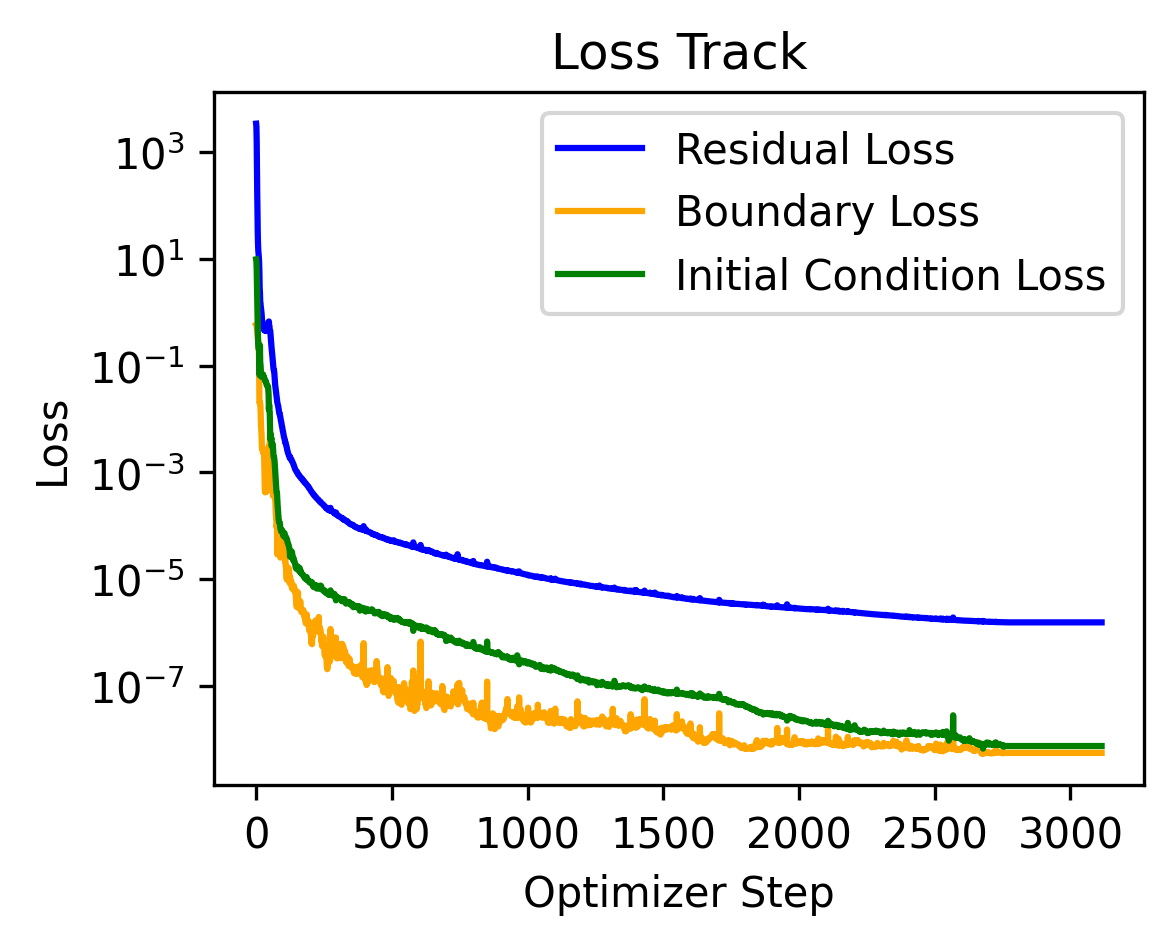}
        \caption{Loss - S-Pformer}
    \end{subfigure}

\end{figure}

\newpage

\subsubsection{1D Wave Loss}

\begin{figure}[H]
    \centering
    \begin{subfigure}[b]{0.45\textwidth}
        \includegraphics[width=\linewidth]{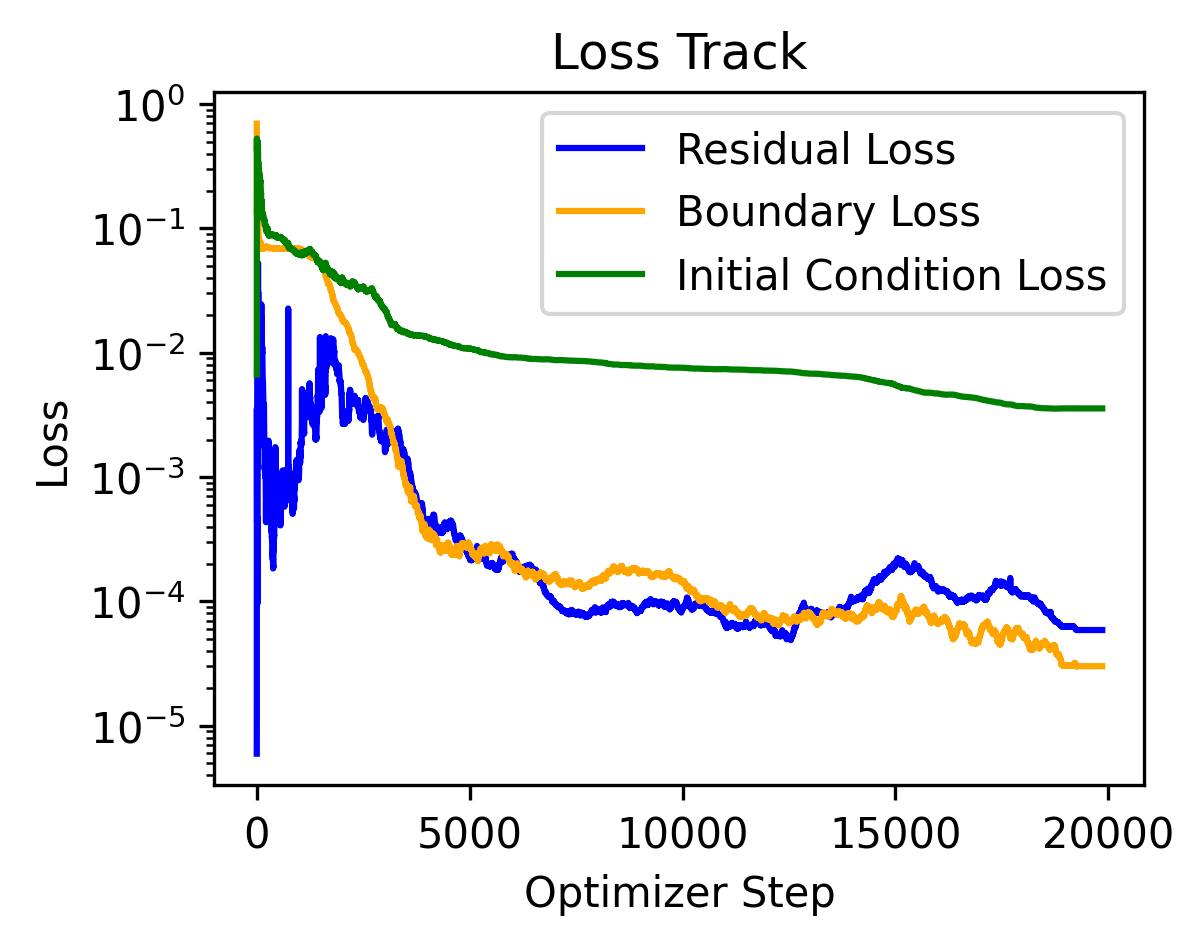}
        \caption{Loss - PINN}
    \end{subfigure}
    \begin{subfigure}[b]{0.45\textwidth}
        \includegraphics[width=\linewidth]{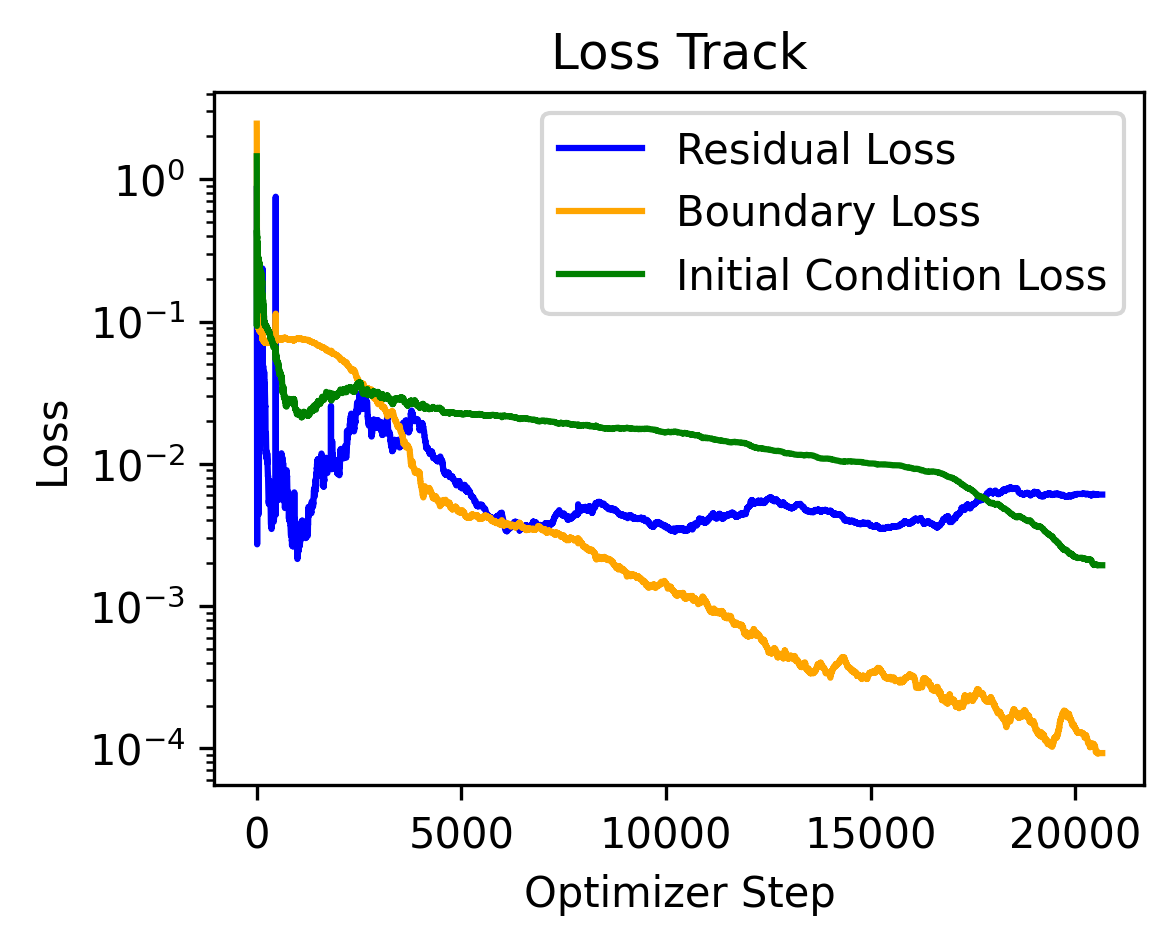}
        \caption{Loss - Pformer}
    \end{subfigure}

\end{figure}

\begin{figure}[H]
    \centering
    \begin{subfigure}[b]{0.45\textwidth}
        \includegraphics[width=\linewidth]{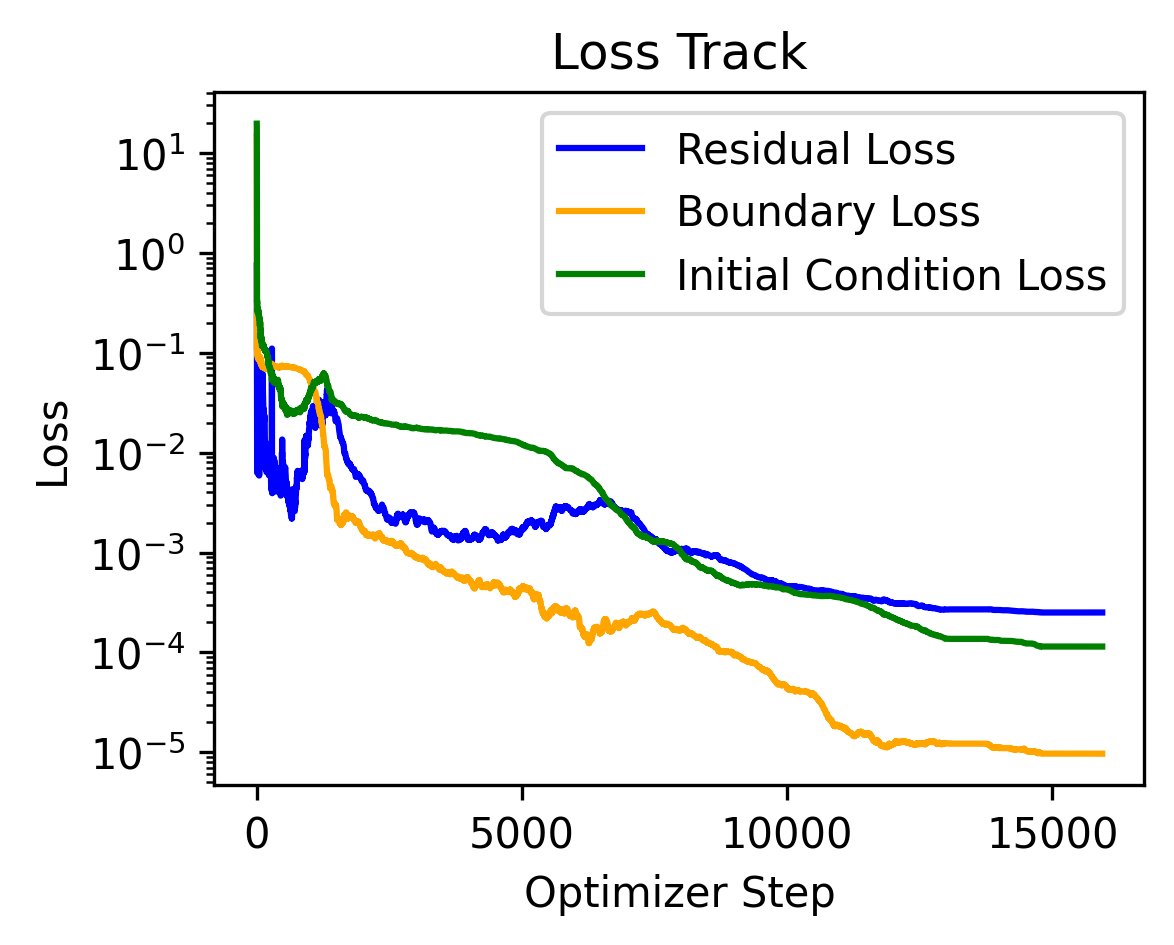}
        \caption{Loss - DO-Pformer}
    \end{subfigure}
    \begin{subfigure}[b]{0.45\textwidth}
        \includegraphics[width=\linewidth]{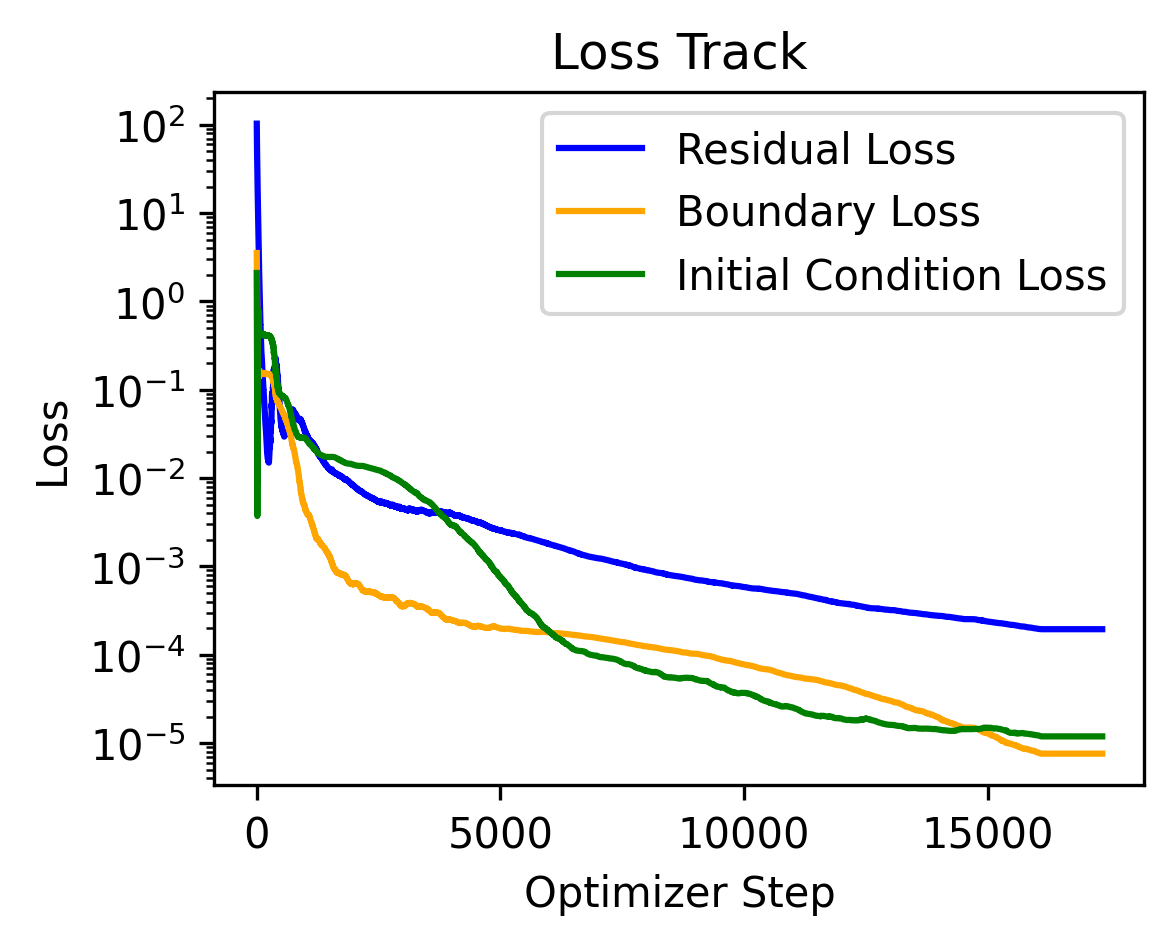}
        \caption{Loss - S-Pformer}
    \end{subfigure}

\end{figure}

\newpage

\subsubsection{2D Navier-Stokes Loss}

Note: There were no initial or boundary conditions on this problem, therefore only the residual loss is plotted. 

\begin{figure}[H]
    \centering
    \begin{subfigure}[b]{0.45\textwidth}
        \includegraphics[width=\linewidth]{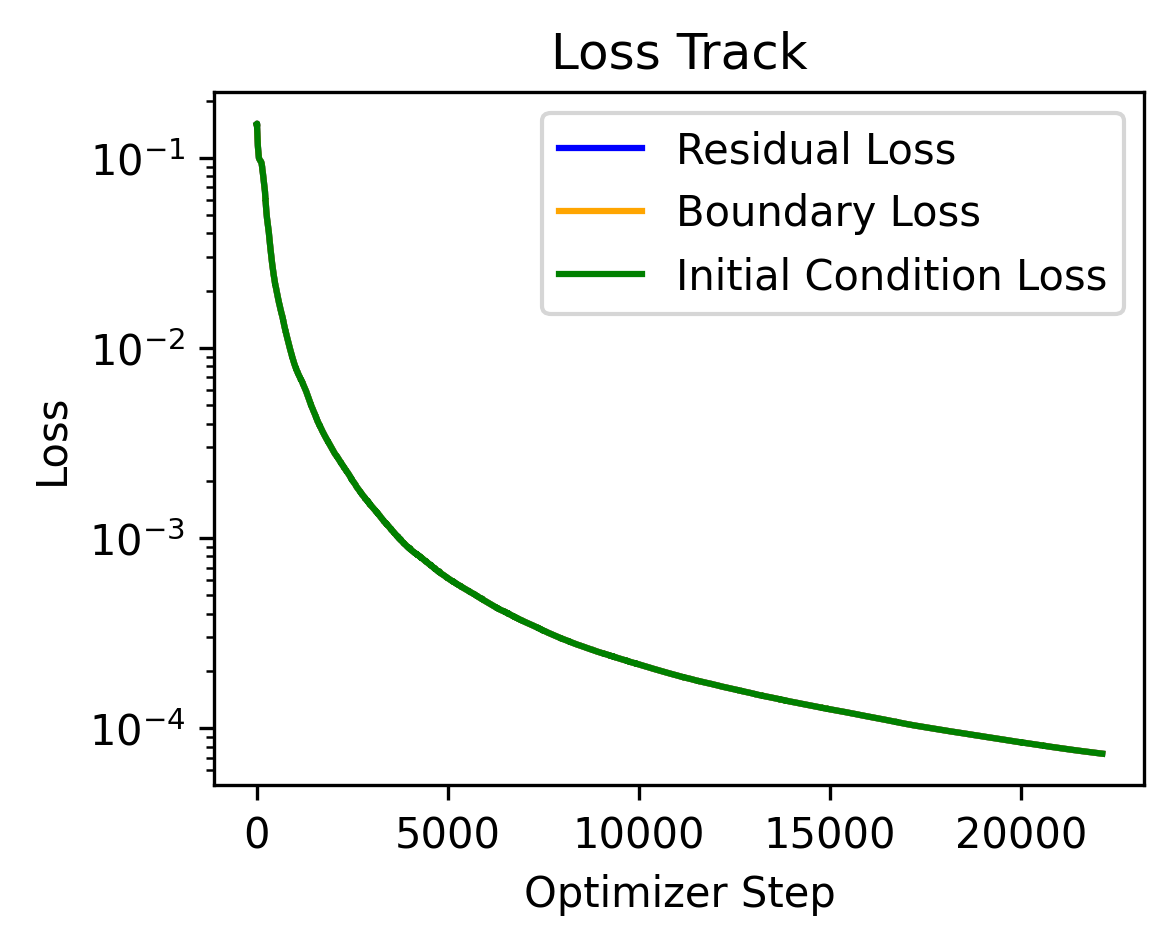}
        \caption{Loss - PINN}
    \end{subfigure}
    \begin{subfigure}[b]{0.45\textwidth}
        \includegraphics[width=\linewidth]{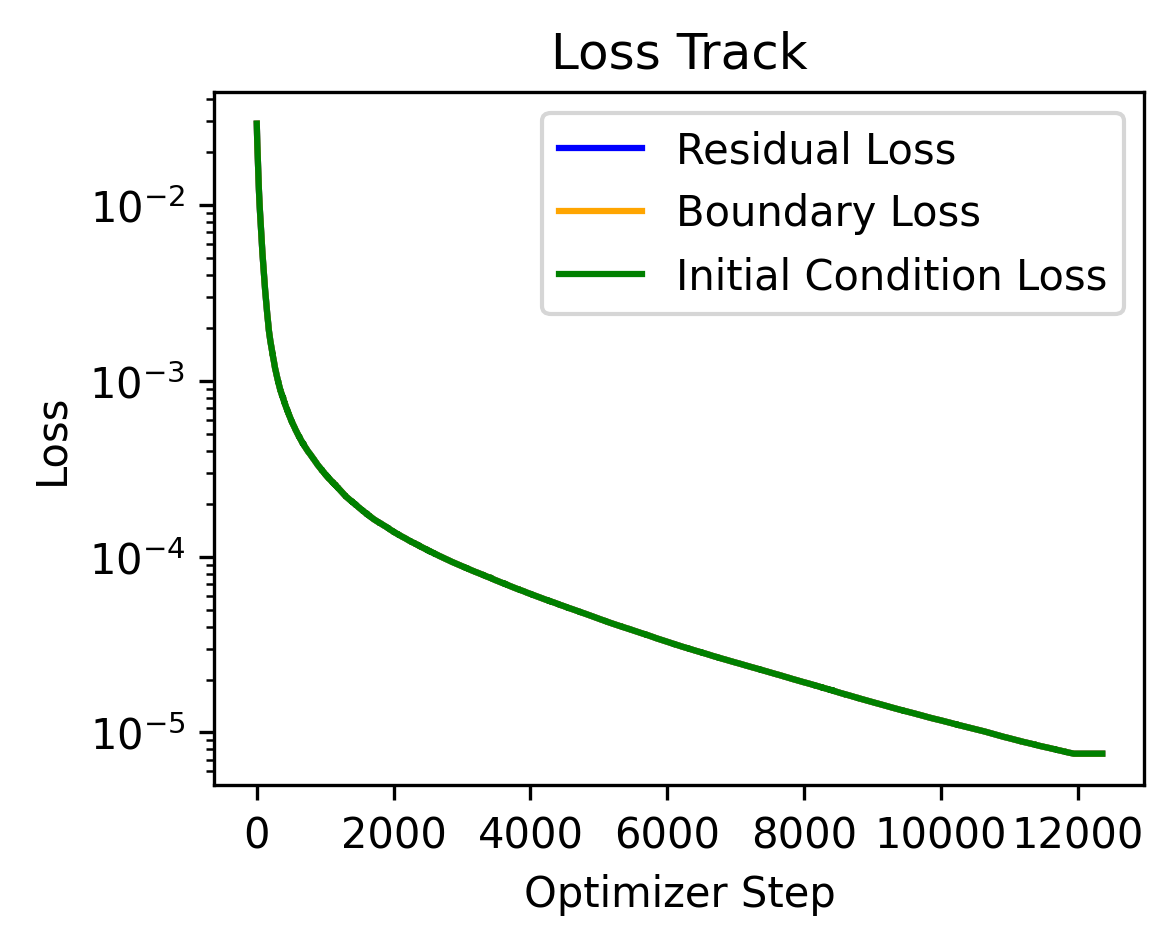}
        \caption{Loss - Pformer}
    \end{subfigure}

\end{figure}

\begin{figure}[H]
    \centering
    \begin{subfigure}[b]{0.45\textwidth}
        \includegraphics[width=\linewidth]{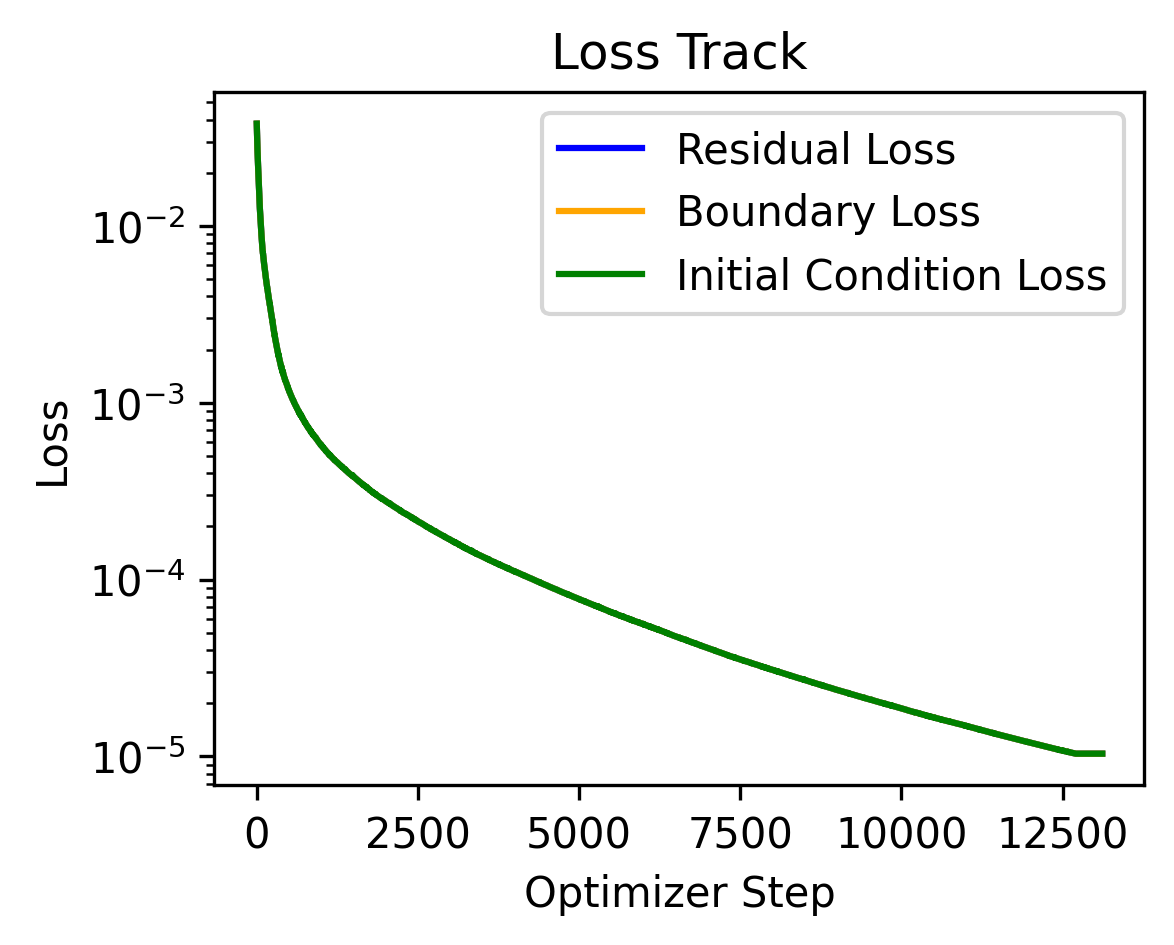}
        \caption{Loss - DO-Pformer}
    \end{subfigure}
    \begin{subfigure}[b]{0.45\textwidth}
        \includegraphics[width=\linewidth]{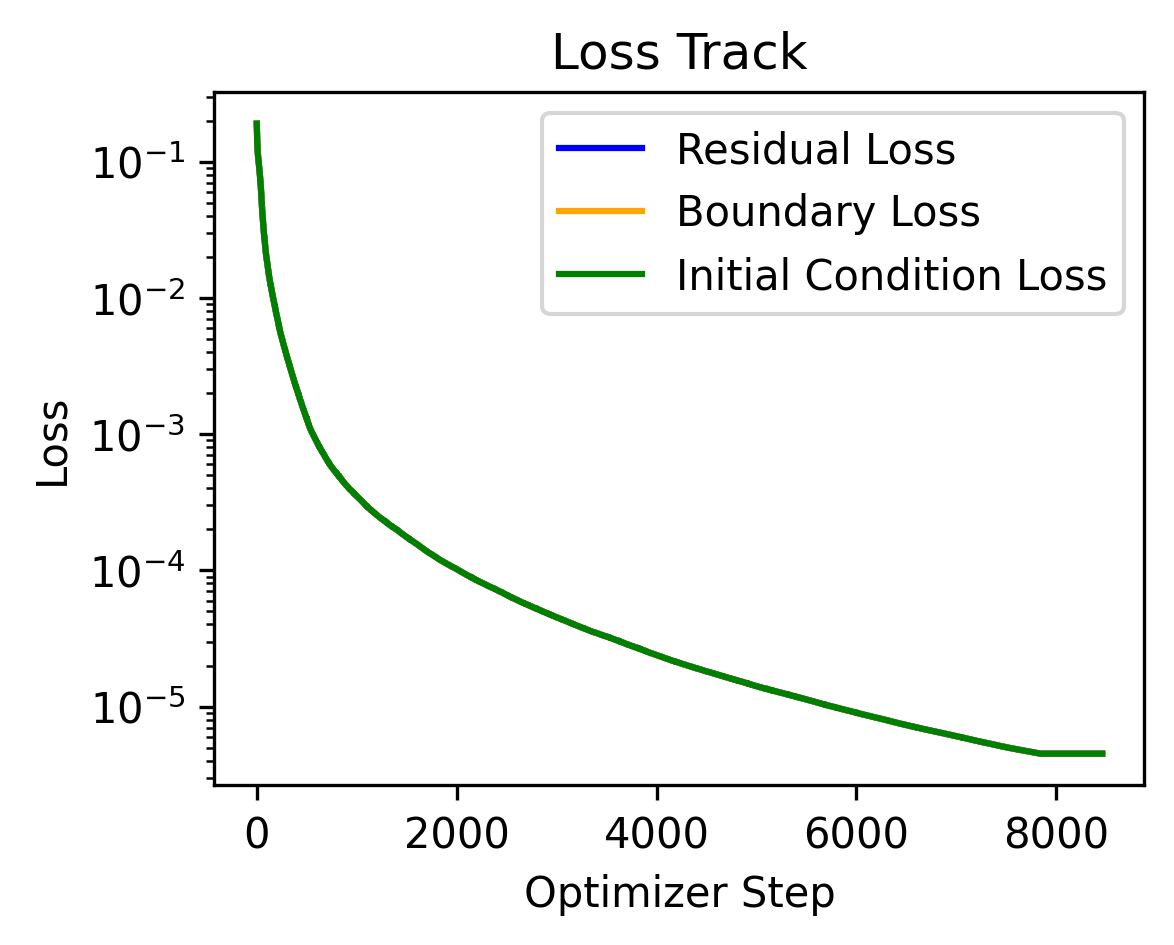}
        \caption{Loss - S-Pformer}
    \end{subfigure}

\end{figure}

\end{document}